\documentclass{article}

\usepackage{microtype}
\usepackage{graphicx}
\usepackage{subfigure}
\usepackage{booktabs} 

\usepackage{hyperref}

\usepackage[accepted]{icml2023}

\usepackage{amsmath}
\usepackage{amssymb}
\usepackage{mathtools}
\usepackage{amsthm}
\usepackage{xspace}
\usepackage{multirow}

\usepackage[capitalize,noabbrev]{cleveref}

\theoremstyle{plain}

\theoremstyle{definition}

\theoremstyle{remark}

\usepackage[textsize=tiny]{todonotes}
\usepackage{lipsum}
\usepackage{afterpage}

\usepackage{wrapfig}

\usepackage{csvsimple}

\definecolor{lightgrey}{rgb}{0.43,0.43,0.43}
\definecolor{darkgreen}{rgb}{0.0,0.7,0.0}

\usepackage{soul}
\definecolor{lightlightgrey}{rgb}{0.9,0.9,0.9}
\sethlcolor{lightlightgrey}

\newcommand{\isat}{ISA-T\xspace}
\newcommand{\isats}{ISA-TS\xspace}
\newcommand{\isatsr}{ISA-TSR\xspace}
\newcommand{\slotpositions}{S_p}
\newcommand{\slotscales}{S_s}
\newcommand{\slotrotm}{S_r}
\newcommand*{\vertbar}{\rule[-1ex]{0.5pt}{2.5ex}}

\newcommand{\ebar}[2]{$#1$ {\color{lightgrey}\tiny $\pm #2$}}

\icmltitlerunning{Invariant Slot Attention}

\begin{document}

\twocolumn[
\icmltitle{Invariant Slot Attention: Object Discovery with Slot-Centric Reference Frames}



\icmlsetsymbol{intern}{*}
\icmlsetsymbol{equal}{$\dagger$}

\begin{icmlauthorlist}
\icmlauthor{Ondrej Biza}{intern,nu}
\icmlauthor{Sjoerd van Steenkiste}{gogl}
\icmlauthor{Mehdi S. M. Sajjadi}{gogl}
\icmlauthor{Gamaleldin F. Elsayed}{equal,gogl}
\icmlauthor{Aravindh Mahendran}{equal,gogl}
\icmlauthor{Thomas Kipf}{equal,gogl}
\end{icmlauthorlist}

\icmlaffiliation{nu}{Northeastern University, Boston, MA, USA.}
\icmlaffiliation{gogl}{Google Research}

\icmlcorrespondingauthor{Ondrej Biza}{biza.o@northeastern.edu}

\icmlkeywords{Spatial symmetry, Equivariance, Abstraction, Object-centric learning, Unsupervised learning}

\vskip 0.3in
]



\printAffiliationsAndNotice{An earlier version appeared at the NeurIPS'22 NeurReps workshop as \href{https://openreview.net/pdf?id=nk_nSogsrZL}{"Spatial Symmetry in Slot Attention"}. $^*$Work done during an internship at Google Research. $^{\dagger}$Equal contribution.} 

\begin{abstract}
Automatically discovering composable abstractions from raw perceptual data is a long-standing challenge in machine learning.
Recent slot-based neural networks that learn about objects in a self-supervised manner have made exciting progress in this direction.
However, they typically fall short at adequately capturing spatial symmetries present in the visual world, which leads to sample inefficiency, such as when entangling object appearance and pose.
In this paper, we present a simple yet highly effective method for incorporating spatial symmetries via slot-centric reference frames. We incorporate equivariance 
to per-object pose transformations into the attention and generation mechanism of Slot Attention by translating, scaling, and rotating position encodings. These changes result in little computational overhead, are easy to implement, and can result in large gains in terms of data efficiency and overall improvements to object discovery.
We evaluate our method on a wide range of synthetic object discovery benchmarks namely Tetrominoes, CLEVRTex, Objects Room and MultiShapeNet, and show promising improvements on the challenging real-world Waymo Open dataset.

\begin{flushleft}
\begin{itemize}
\item[+] JAX/FLAX source code: \href{https://github.com/google-research/google-research/tree/master/invariant_slot_attention}{https://github.com/google-research/google-research/tree/master/invariant\_slot\_attention}
\item[+] Model checkpoints: \href{https://huggingface.co/ondrejbiza/isa}{https://huggingface.co/ondrejbiza/isa}
\item[+] Interactive demo: \href{https://huggingface.co/spaces/ondrejbiza/isa}{https://huggingface.co/spaces/ondrejbiza/isa}
\end{itemize}
\end{flushleft}

\end{abstract}

\section{Introduction}
\label{sec:intro}

\begin{figure}
    \centering
    \includegraphics[width=0.9\linewidth]{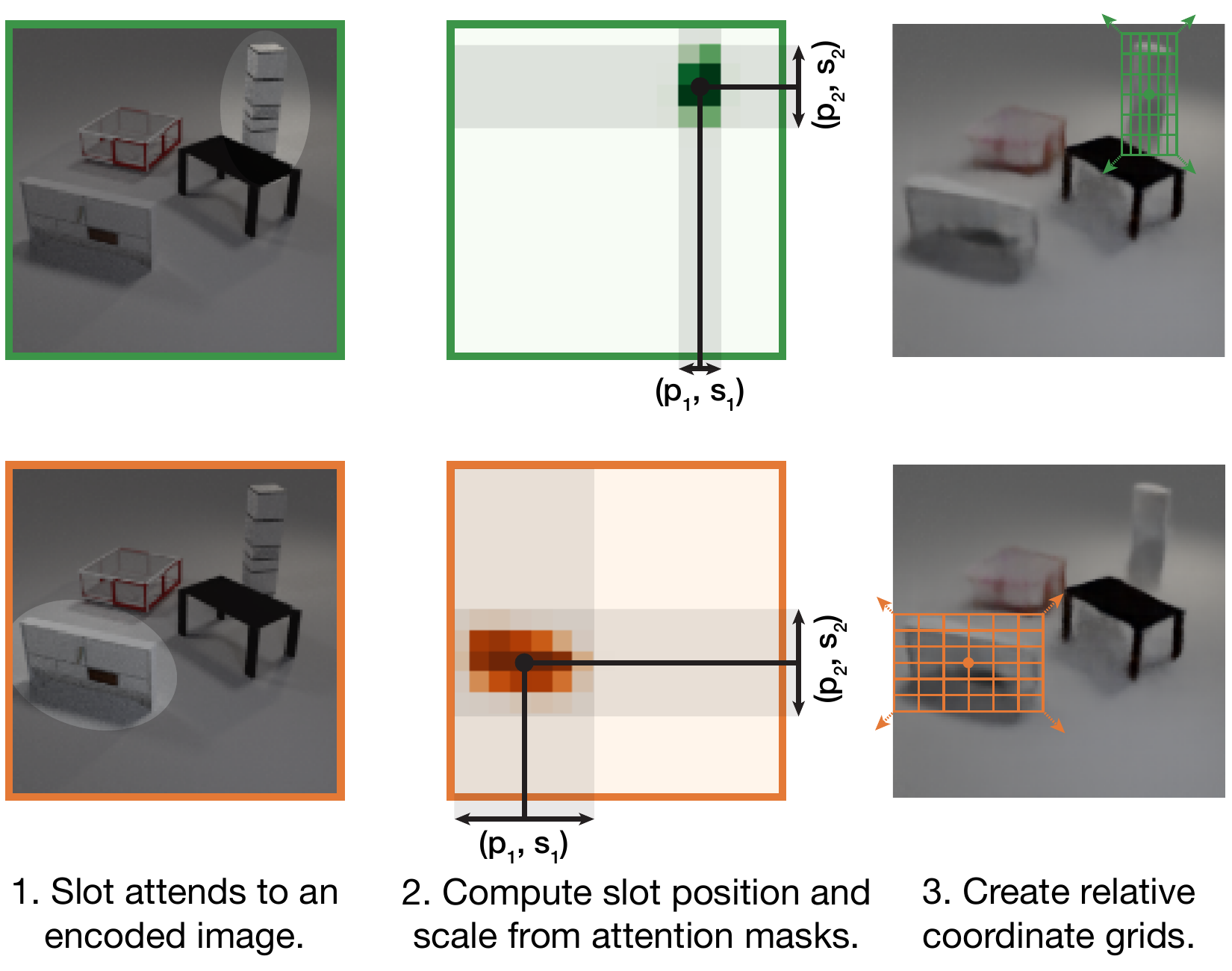}
    \vspace{-1em}
    \caption{Left: Input image. Middle: Invariant Slot Attention masks. Right: Decoded images with relative coordinate grids (limited to $6{\times}6$ grids for ease of visualization).}
    \vspace{-.5em}
    \label{fig:main}
\end{figure}

\looseness=-1Humans have the intrinsic ability to form high level abstractions of objects in visual scenes regardless of their 
appearance \cite{barlow09grandmother}.
In the neuroscience literature, it is hypothesized that humans place canonical `reference frames' onto objects~\citep{bottini2020knowledge}. Such reference frames entail a coordinate system around each object that may allow for forming robust object representations in a viewpoint-invariant manner \cite{hinton81parallel}, and thus for making future predictions by manipulating those reference frames. In a scene with several objects and parts, \citet{hawking19framework} posit that human grid cells represent a hierarchy of reference frames (e.g. the frame of a person's hand, the frame of the mug it is holding, the frame of the mug's handle) and hypothesize that there are special grid cells that model relative transformations between different levels in the scene.

\looseness=-1In machine learning, it is also understood that such \emph{invariant} object representations could be similarly beneficial.
In order to learn per-object invariances, however, one needs robust models that can decompose visual scenes into objects. This topic, known as object-centric representation learning, has proven to be challenging in the absence of direct supervision~\cite{greff17neural}. Yet, in the past few years, powered by more complex datasets, scalable architectures, and improved compute infrastructure, substantial progress has been made in discovering object representations from raw perceptual data.
This includes the use of (inverted) dot-product attention~\cite{locatello20object}, powerful decoders~\cite{singh21illiterate,singh22simple}, optical flow~\cite{kipf22conditional,xie2022segmenting}, depth maps~\cite{elsayed22savipp}, pre-trained features~\cite{seitzer23bridging}, etc.
These advances have made an investigation into object-level invariances and their benefits for object representation learning more tractable. 

\begin{figure}
    \centering
    \includegraphics[width=\linewidth]{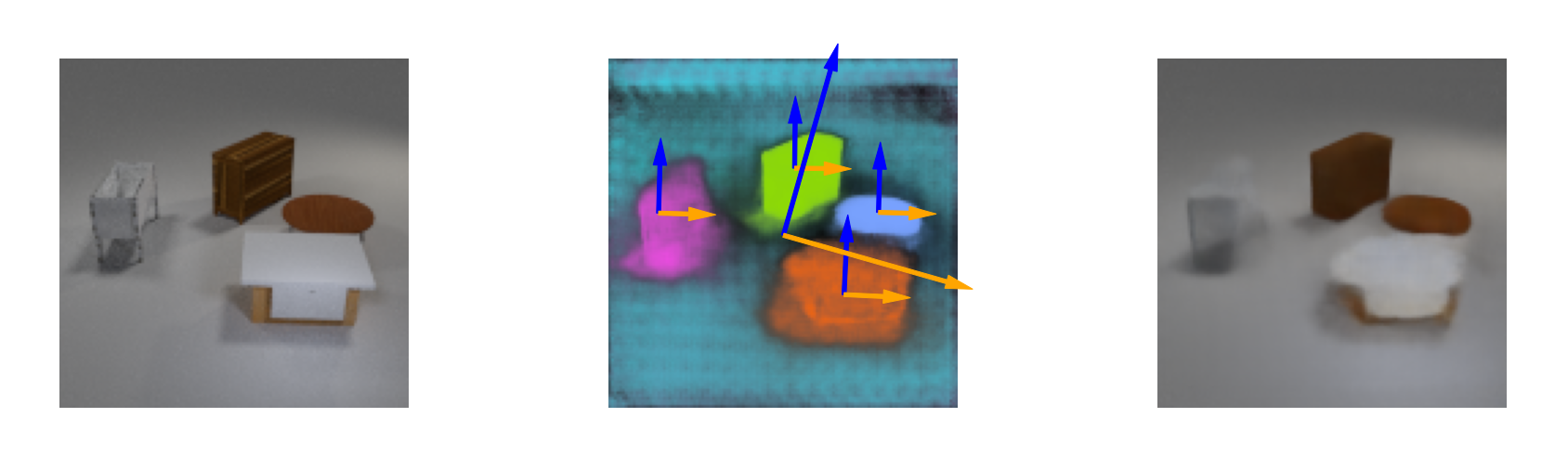}
    \includegraphics[width=\linewidth]{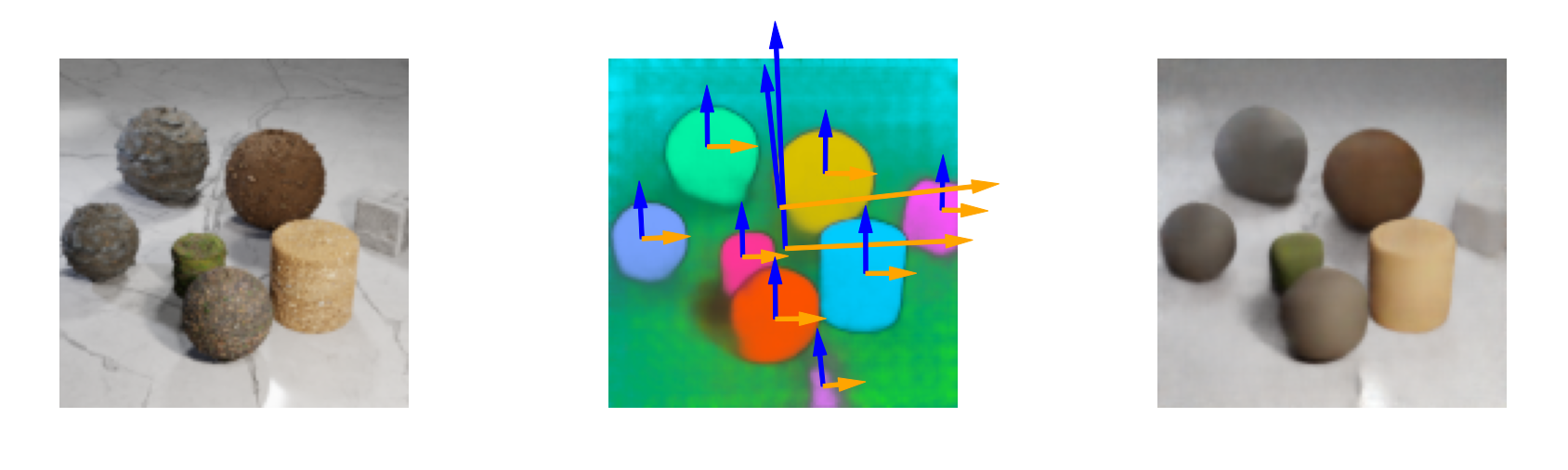}
    \vspace{-2em}
    \caption{Examples of reference frames learned by Invariant Slot Attention. From left to right: input image, soft segmentation masks with a pair of arrows for each slot, decoded image.}
    \label{fig:frames}
\end{figure}

Building upon these advances, and inspired by prior work such as AIR \cite{eslami16attend} and Spatial Transformers \cite{jaderberg15spatial}, invariance to object pose can be established by assigning a reference frame to each object. Any percept related to the object (such as spatial features in a feature map) can then be processed in a coordinate system relative to the object's reference frame, and any prediction about the object (such as a reconstruction of its appearance) retains the pose invariance property. Figure \ref{fig:main} shows an example of such a system in the setting of self-supervised object discovery. Figure \ref{fig:frames} shows examples of learned reference frames.

Reasoning in relative reference frames is an example of a spatial symmetry; machine learning models can leverage these symmetries to improve sample-efficiency, generalization and prediction consistency \citep{bronstein21geometric}. Much work has been done in leveraging \textit{exact} spatial symmetries, for example in protein dynamics modelling \citep{han22equivariant} or in constrained robotics environments \citep{wang20policy}. Yet, these symmetries have been explored only to a limited extent in scene understanding, where an object's appearance might change due to lighting or occlusions, and the effect of 3D rotation on object appearance cannot be easily formalized \citep{park22learning}.

We focus our paper on the topic of unsupervised object discovery with slot-based models \cite{greff19multiobject,greff2020binding}. Slot-based models compress a scene into a discrete number of latent variables--``slots''. Slots can learn to represent individual objects in the scene without additional supervision. In these models slots are assumed to be symmetric under permutation, up to the choice of initialization, which can facilitate object discovery.
Yet, other spatial symmetries have been explored only to a limited extent. While earlier methods \citep{jaderberg15spatial,eslami16attend} use explicit per-object poses in the decoder, their monolithic encoders still operate in terms of absolute scene coordinates and thus do not fully account for spatial symmetries.

In our work, we draw a connection between spatial symmetries and slot-based models via the use of \textit{pose-relative position encoding}. We base our work on the Slot Attention~\citep{locatello20object} architecture, which is a widely used slot-based model for object discovery. Slot Attention uses position encoding to map from the grid-structured image representation to its set of latent slots via an attention mechanism, and similarly uses position encoding to map back into the image space in the decoder. Our core insight is that we can achieve equivariance to spatial symmetry transformations (i.e., pose transformations) at a per-object level by transforming the respective position encodings relative to the each object's inferred pose. This presents an elegant yet computationally efficient solution for integrating spatial symmetries in object-centric architectures.

In our experiments, we demonstrate that incorporating translation and scale symmetries into Slot Attention results in improved data efficiency, out-of-domain generalization, and often overall improvements in object discovery performance with little computational overhead. While our framework also allows for incorporation of approximate 2D rotation symmetry, we find that this frequently only leads to minor (or no) benefits, likely due to the nature of our datasets (images of 3D scenes). We evaluate our model, Invariant Slot Attention (ISA), on standard synthetic object discovery benchmarks, a challenging textured dataset, and in object discovery in a real-world autonomous driving dataset.

\section{Related Work}
\label{sec:related_work}


\textbf{Unsupervised Object Discovery.}\quad
Object discovery using model inductive biases has been an active and growing area of research. Early works use an iterative inference setting: a model alternates between reconstructing an image and inferring the membership of pixels into groups forming object representation vectors (or \textit{slots}) \cite{greff15binding,greff16tagger,greff17neural,steenkiste18relational,greff19multiobject}. These works do not implement spatial symmetries outside of the use of convolutional networks. \citet{burgess19monet,engelcke20genesis} also fall in this category, wherein they auto-regressively predict per-slot attention masks, which are used to compute unstructured slot latent vectors.

\looseness=-1In a separate line of work, \citet{eslami16attend,kosiorek18sequential,jiang20scalor} use a monolithic encoder to predict the position and scale of each object ($z_{\text{where}}$). A Spatial Transformer (ST) \cite{jaderberg15spatial} decodes object appearance invariant to $z_{\text{where}}$. To enable decoding of complex images, \citet{monnier21unsupervised,smirnov21marionette,sauvalle22unsupervised} combine ST decoders with background prediction models. During encoding, partial translation equivariance can be achieved by predicting object positions and scales relative to anchors in a grid \cite{crawford19spatially,lin20space,jiang20generative}.

Our work builds on the Slot Attention model (Section \ref{sec:background}), which groups pixels into slots by iteratively computing cross-attention between inputs and slot latent vectors \cite{locatello20object}. We take inspiration from capsules \cite{hinton11transforming,sabour17dynamic,hinton18matrix,kosiorek19stacked}, which model the hierarchy of relative transformations between object parts forming objects (and objects forming a scene). \citet{hinton21glom} points out that slots in Slot Attention can be thought of as \textit{universal} capsules that learn to detect any object in any position and orientation. Unlike capsules, Slot Attention does not natively model the part-whole hierarchy. Parts of an object (spatial features from a CNN encoder) are anchored in the reference frame of the camera, instead of being represented relative to the position, orientation and scale of the object. Hence, the original Slot Attention model is not predisposed to spatially invariant reasoning about objects and their parts. Since Slot Attention has been the basis of several recent follow-up works~\cite{emami21efficient,singh21illiterate,sauvalle22unsupervised}, video modeling \cite{kipf22conditional,elsayed22savipp,wu22slotformer}, multi-view scene understanding \cite{sajjadi21scene}, and panoptic video segmentation~\cite{zhou2021slot}, to name a few, making it spatially invariant has the potential for broader impact.

\textbf{Relative Position Encoding.}\quad
Follow-up work to the original Transformer model \cite{vaswani17attention}, replaces absolute position encodings with an encoding of the relative distance between a pair of words \cite{shaw18selfattention}. Since word or character distances are discrete, most works associate a learnable vector with each possible relative distance between words. Similarly, relative position encoding has been explored in the context of Vision Tranformers~\cite{dosovitskiy21image,cordonnier2020relationship}: both in terms of discrete~\citep{bello19attention,parmar19standalone,wang20axialdeeplab,srinivas21bottleneck} as well as continuous encodings between patches~\citep{zhao20exploring}. We refer to \citet{dufter22position} for a recent review. These methods consider relative position encoding in an encoder-only setup, which is orthogonal to our pose-relative position encoding mechanism that operates on cross-attention and related mappings between visual tokens and object slots.

Ideas for relative position encoding in vision have further been explored in the context of the Detection Transformer (DETR)~\cite{carion20endtoend}.
\citet{zhu21deformable} take a step towards disentangling object query positions in DETR for supervised detection tasks: attention is only computed over the local neighborhood of input tokens close to a reference point. Similarly, \citet{meng21conditional,gao21fast} modulate DETR's decoder attention by predicting the supposed center (and possibly scale) of each detected object from object queries in each decoder layer. Finally, \citet{wang22anchor,liu22dabdetr,zhang22dino} show that  separation of object positions and scales from object appearance leads to significant improvements. Unlike our method, their approach still operates on absolute object coordinates either as explicit input or output of learnable modules, i.e. it does not fully respect spatial symmetries. In ISA, we extract positions and scale from cross-attention masks, which is expected to ensure spatial equivariance in our architecture.


\section{Background: Slot Attention}
\label{sec:background}

\newcommand{\sizec}{130px}
\begin{figure*}[t!]
    \centering
    \includegraphics[width=0.9\textwidth]{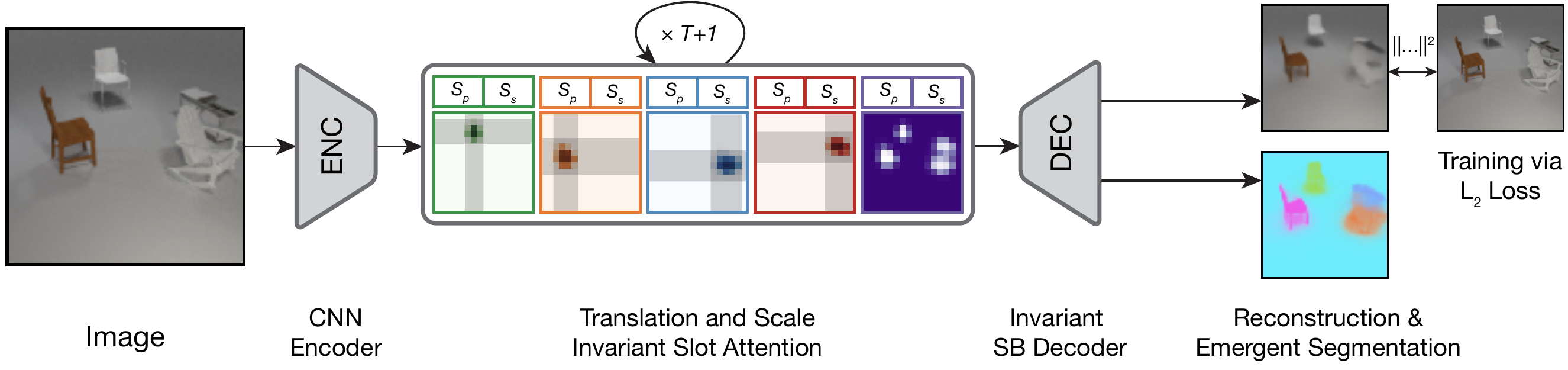}
    \vspace{-1em}
    \caption{Invariant Slot Attention (ISA-TS) with an input image from MultiShapeNet (left), visualization of five intermediate attention masks (top; the purple masks corresponds to a background slot), and the decoded image and segmentation mask (right).}
    \label{fig:method}
\end{figure*}

Our method for object discovery with slot-centric reference frames is based on the Slot Attention~\citep{locatello20object} architecture, which presents a simple yet effective attention-based method for decomposing scenes into objects and for learning object representations from un-annotated image data. Slot Attention consists of an encoder, an attention mechanism, and a decoder.

The encoder ($E_{\theta}$), e.g.~a convolutional neural network (CNN), maps images into an intermediate representation of $N=H' \times W'$ tokens of dimension $D_t$. The attention mechanism in Slot Attention (SA) simulates a competition of latent slots ($\mathrm{slots} \in \mathbb{R}^{K{\times}D_s}$) over input tokens ($\mathrm{inputs} \in \mathbb{R}^{N{\times}D_t}$). Intuitively, slots win tokens if they are able to reconstruct the corresponding parts of the input image -- these parts usually correspond to entire objects or coherent object parts.

The model computes 
a form of 
cross-attention \cite{luong15effective,vaswani17attention} between $\mathrm{slots}$, transformed into $\mathrm{queries} = \mathcal{Q}(\mathrm{slots})$, 
and $\mathrm{inputs}$, transformed into $\mathrm{keys}$ and $\mathrm{values}$:
\begin{align}
    \mathrm{keys} &= f\left(\mathcal{K}(\mathrm{inputs}) + g(\mathrm{abs\_grid})\right)\label{eqn:absk}\,\\
    \mathrm{values} &= f\left(\mathcal{V}(\mathrm{inputs}) + g(\mathrm{abs\_grid})\right)\label{eqn:absv}\,
\end{align}
%
Here $\mathrm{abs\_grid}$ encodes the absolute positions of input tokens in an image, $\mathcal{K},\mathcal{V},\mathcal{Q},g$ are linear functions and $f$ is an MLP. The position encodings are horizontal and vertical coordinate grids scaled to [-1, 1]. The dot product between $\mathrm{queries}$ and $\mathrm{keys}$ yields the attention weights, which are then normalized over $\mathrm{queries}$ to facilitate competition between slots. The output of cross-attention is computed in several iterations; in each iteration, a recurrent neural network updates the slot representations (App. Algorithm \ref{alg:isa}).

The decoder ($D_{\phi}$) reconstructs the input image based on the slot latent vectors. We use the Spatial Broadcast (SB) decoder \cite{watters19spatial}, which first repeats each slot $H'*W'$ times to create the initial spatial feature map, adds position encoding as in the encoder, and then maps it to (R,G,B,$\alpha$) values for each pixel. 
The $\alpha$ mask, which represents the predicted segmentation mask, is normalized over slots. The final image is created by taking the weighted sum over $K$ slots for each pixel. 

\section{Invariant Slot Attention}
\label{sec:method}

In Invariant Slot Attention (ISA), Figure~\ref{fig:method}, we introduce mechanisms for invariant encoding and decoding to the Slot Attention architecture, with the goal to disentangle slot appearance from slot position, orientation and scale. 

In our formulation, slot latent vectors are \emph{invariant} to object pose, while slot position, orientation and scale are \emph{equivariant} to pose transformations. Note that slot scale refers to the spatial extent in pixel space and therefore accounts for both the object's size as well as distance to the camera. 
Our mechanism can be implemented in two easy steps:

\textbf{Step 1:} Either initialize slots with explicit reference frames in addition to their latent vectors that encode appearance, or, if available, compute slot reference frames from per-slot attention masks.

\textbf{Step 2:} Create relative position encodings, $\mathrm{rel\_grid}$, in each iteration of Slot Attention and in the SB decoder using reference frames (slot poses) from step 1.

The following subsections explain the above steps in detail along with how $\mathrm{rel\_grid}$ can be used instead of $\mathrm{abs\_grid}$ in Slot Attention iterations and the SB decoder.

\subsection{Translation and Scaling Invariant Slot Attention}\label{sec:isa}

To achieve invariance to translation and scaling, we instantiate 2D slot positions ($\slotpositions$) and scales ($\slotscales$), which can be initially sampled randomly or learned (Appendix \ref{appendix:model}). $\left(\slotpositions, \slotscales\right)$ define a reference frame for each slot. We use it to translate and scale the input position encoding for each slot $k\in\{1,...,K\}$ separately:
\begin{align}
\mathrm{rel\_grid}^k = (\mathrm{abs\_grid} - \slotpositions^k)\ /\ \slotscales^k
    \label{eq:relative_grid}
\end{align}
We translate and scale by the \textit{inverse} of slot positions and scales. Intuitively, we are undoing the estimated object pose so that we can process it in a canonical reference frame. To compute attention masks, we create per-slot keys and values with $\mathrm{rel\_grid}$ instead of $\mathrm{abs\_grid}$. That is, $\forall\,1 \leq k \leq K$:
\begin{align}
    \mathrm{keys}^k &= f\left(\mathcal{K}(\mathrm{inputs}) + g(\mathrm{rel\_grid}^k)\right)\, \\
    \mathrm{values}^k &= f\left(\mathcal{V}(\mathrm{inputs}) + g(\mathrm{rel\_grid}^k)\right)\,
\end{align}
Note that, compared to \cref{eqn:absk,eqn:absv}, we now have $N*K$ keys and values, and $K$ queries. The original Slot Attention has only $N$ keys and values. We find the increase in computational cost negligible in the standard regime of detecting around 10 objects in an image.

To infer $S_p$ and $S_s$, we use the obtained per-slot keys and values to compute attention weights, $\mathrm{attn}$, from which we in turn extract new slot positions and scales:
\begin{align}
    &\slotpositions = \frac{\sum_{n=1}^{N} \mathrm{attn}_n * \mathrm{abs\_grid}_n}{\sum_{n=1}^{N} \mathrm{attn}_n}\,\\
    &\slotscales = \sqrt{\frac{\sum_{n=1}^{N} (\mathrm{attn}_n + \epsilon) * (\mathrm{abs\_grid}_n - \slotpositions)^2}{\sum_{n=1}^{N} (\mathrm{attn}_n + \epsilon)}}\,
\end{align}
The intuition behind this process is that attention weights focus on the relevant object in the image, and we can use the center of mass of the attention mask to infer the objects' position, and its spread to infer its scale.

The same process is repeated in each Slot Attention iteration. Afterward, we run one additional iteration to compute slot statistics without updating the slot latent vectors.

In the Invariant Spatial Broadcast decoder, we create pose-relative position encoding using the same process as in the encoder. We use the final per-slot reference frames $\left(\slotpositions,\slotscales\right)$ computed in Slot Attention and we compute $\mathrm{rel\_grid}$ as in eqn.~(\ref{eq:relative_grid}). $\mathrm{rel\_grid}$ is then projected to $D_s$, the slot dimension, using a linear function $h$ and added to spatially-broadcasted slots $\mathrm{SB} \in {\mathbb{R}^{K{\times}H'{\times}W'{\times}D_s}}$:
\begin{equation}
    (R,G,B,\alpha) = D_{\phi}(\mathrm{SB} + h(\mathrm{rel\_grid}))
\end{equation}
    where $\mathrm{rel\_grid} \in {\mathbb{R}^{K{\times}H'{\times}W'{\times}2}}$.
All changes described in this section can be implemented in a few lines of code (Appendix Algorithm \ref{alg:isa}). Note that we backpropagate through both the computed slot positions and scales, and the relative coordinate grids.

Although we demonstrate our idea with image-based Slot Attention and 2D spatial symmetries, the same principle could apply in videos, by processing each frame with the proposed method, \cite{kipf22conditional,elsayed22savipp}, or in 3D with slot-based Neural Radiance Fields \cite{mildenhall20nerf} or Object Scene Representation Transformers \cite{sajjadi22object}, by equipping slots with positions extracted from multiple views, \cite{stelzner21decomposing,sajjadi21scene,yu22unsupervised}.

\subsection{Invariance to Rotations}

The orientation of an object is much more ambiguous than its position and scale. To study this symmetry, we use a simple principal component heuristic \cite{yi00principal} to estimate slot poses. We estimate the orientation of an object using the axis with the highest variation (the first principal component) of the attention mask. We then construct a rotation matrix $\slotrotm$ with the principal and an orthogonal axis, with rotations limited to $\pi/4$. 
Further details are provided in Appendix~\ref{appendix:model:rot}.
We then compute the relative position encoding by inverse translation, rotation and scaling:
\begin{align}
    &\mathrm{rel\_grid}^k = (S^k_r)^{-1} (\mathrm{abs\_grid} - \slotpositions^k)\ /\ \slotscales^k\,.
    \label{eq:relative_grid:rot}
\end{align}

\section{Experiments}
\label{sec:exp}

\newcommand{\sizeb}{41.5pt}
\begin{figure}[h!]
    \centering
    \includegraphics[height=\sizeb]{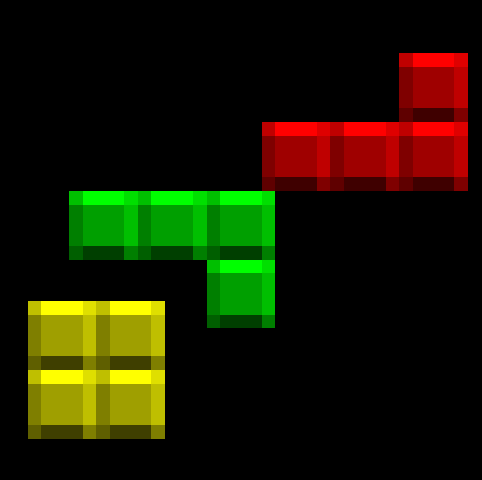}~%
    \includegraphics[height=\sizeb]{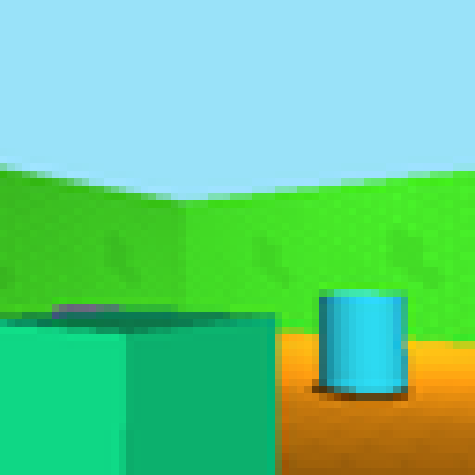}~%
    \includegraphics[height=\sizeb]{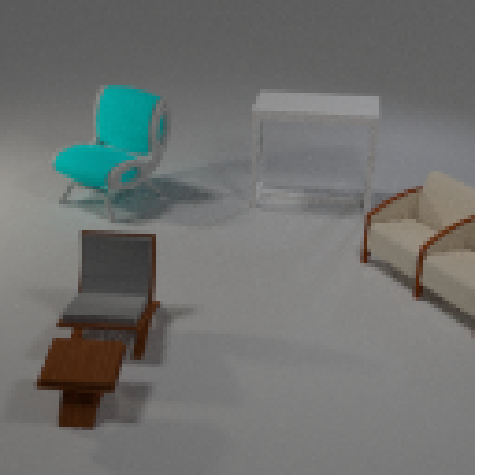}~%
    \includegraphics[height=\sizeb]{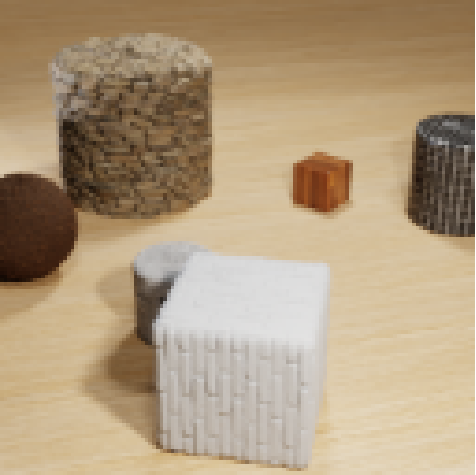}~%
    \includegraphics[height=\sizeb]{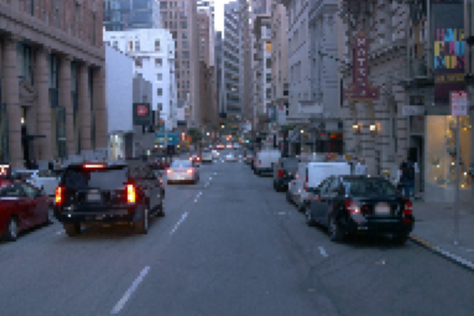}
    \vspace{-1.5em}
    \caption{\textbf{Datasets} (left to right):  Tetrominoes, Objects Room, MultiShapeNet, CLEVRTex, and Waymo Open.}
    \label{fig:dataset_examples}
\end{figure}

We evaluate Invariant Slot Attention (ISA) on six unsupervised object discovery datasets illustrated in fig. \ref{fig:dataset_examples}, which include both synthetic and real-world datasets. We measure the Adjusted Rand Index for foreground object segments (FG-ARI) and the mean squared error of decoded images (MSE) summed over all pixels and averaged over images, please see Appendix \ref{appendix:metrics} for details.

\subsection{Proof of concept: Tetrominoes}
\label{sec:exp:tetro}

\newcommand{\sizea}{92pt}
\begin{figure*}
    \centering
    \subfigure[Sample efficiency.\label{fig:tetro:sample}]{\includegraphics[height=\sizea]{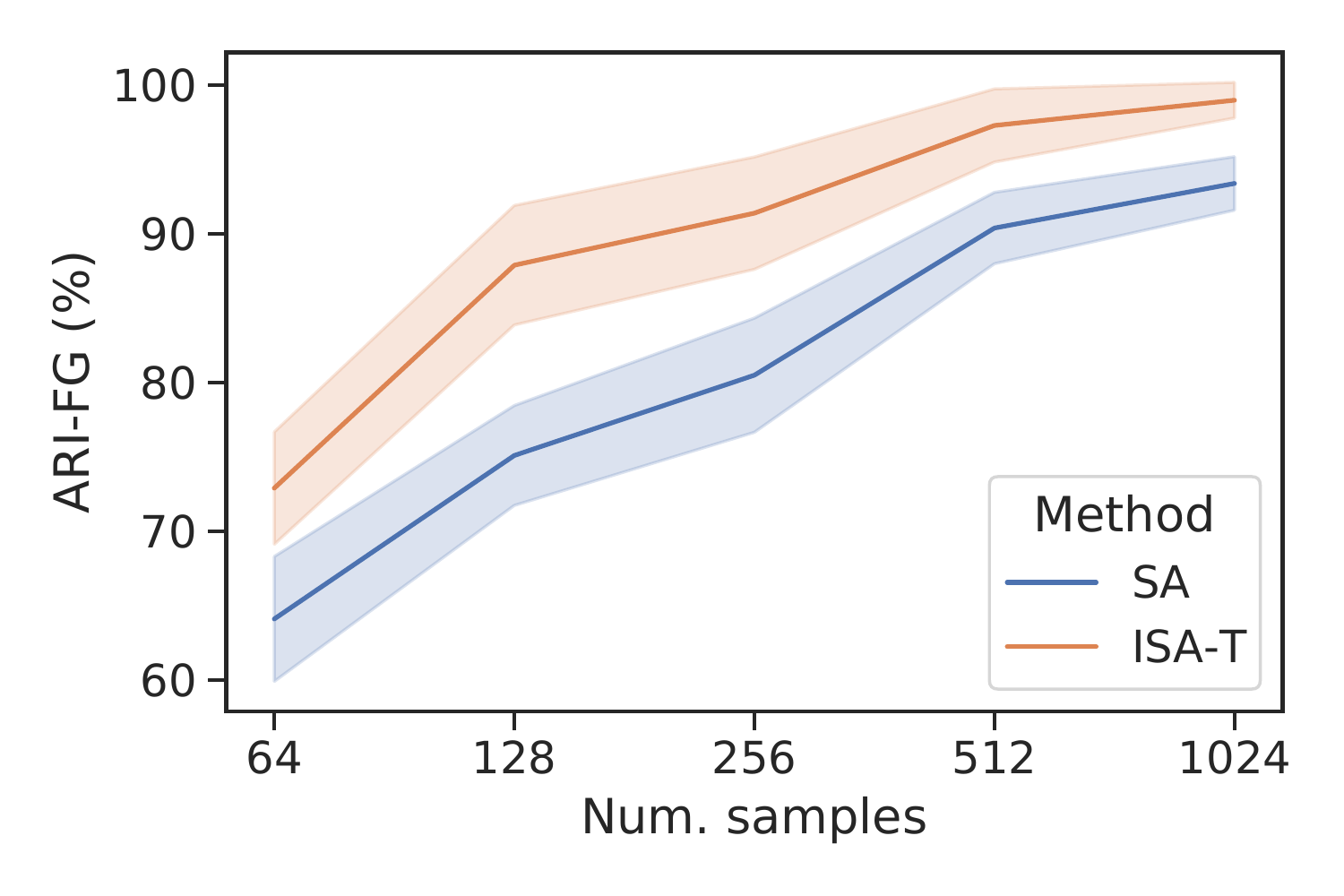}}
    \subfigure[Test-time OOD generalization. From left-to-right: The setup, non-invariant baseline output, ISA-T (our) output, and FG-ARI for both methods (10 seeds per method).\label{fig:tetro:left_right}]{
        \includegraphics[height=\sizea]{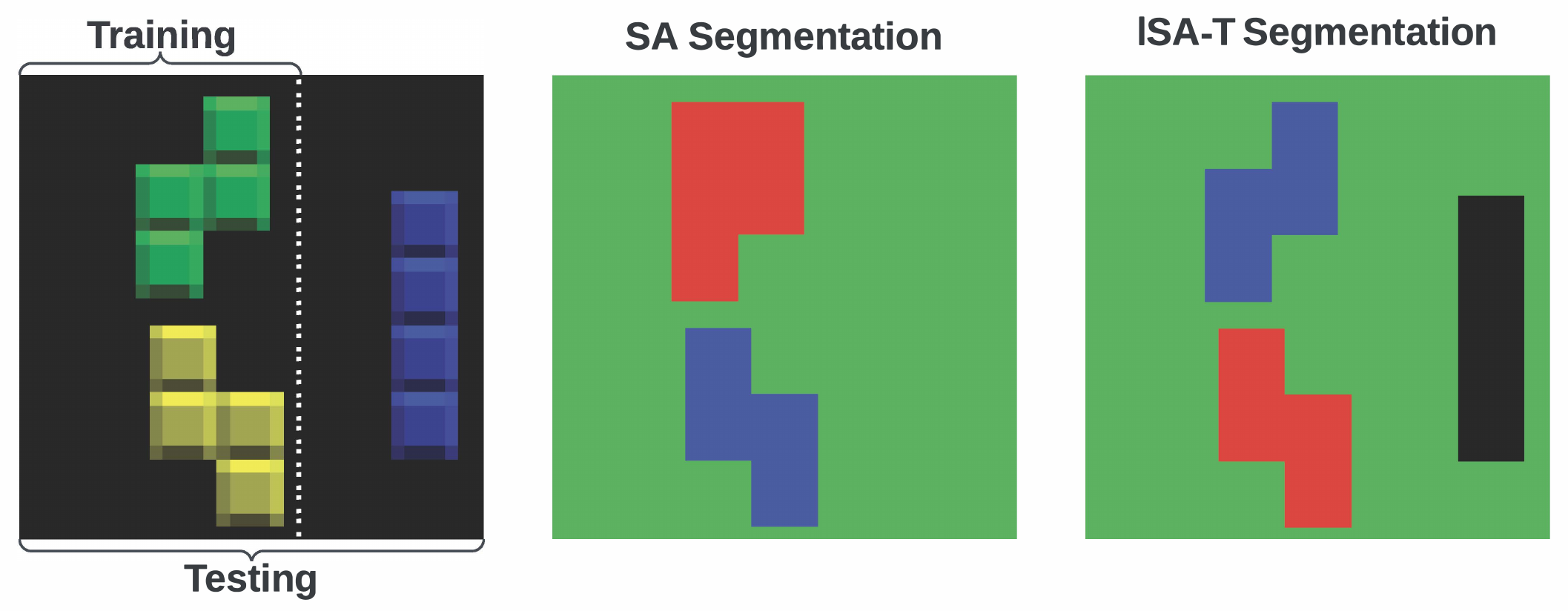}
        \includegraphics[height=\sizea]{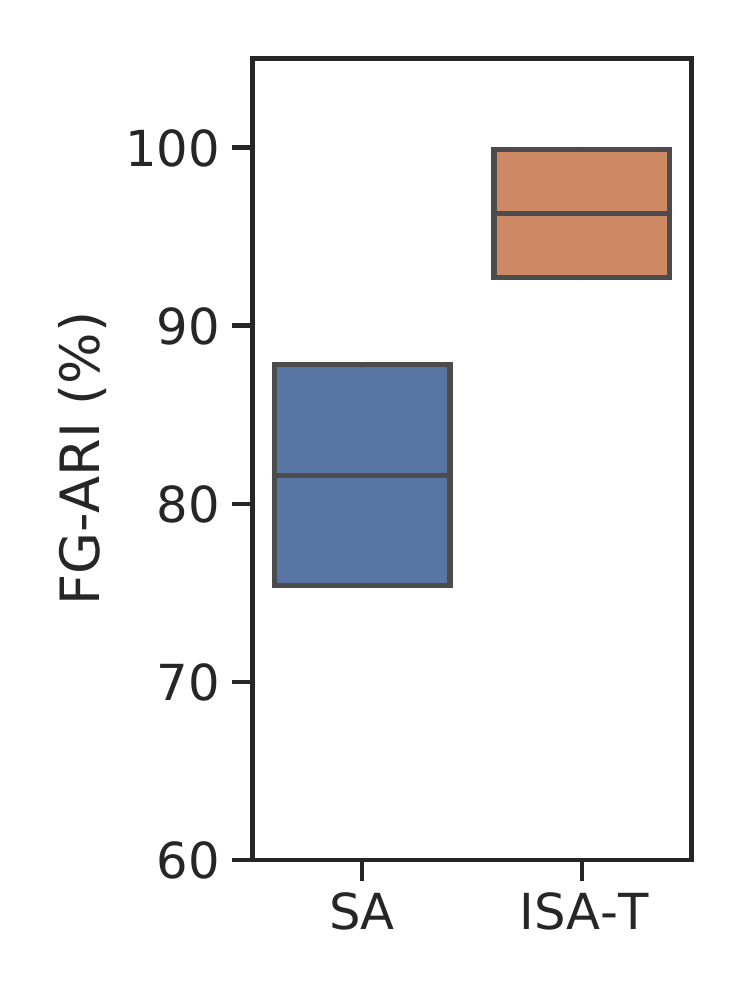}
    }
    \caption{Comparing between Slot Attention (SA) and Translation Invariant Slot Attention (\isat) on Tetrominoes. \isat (a) reaches the same performance as SA with 2 to 4 times fewer samples, and (b) is able to generalize to OOD object positions.}
    \label{fig:tetro}
\end{figure*}

Starting with a proof of concept, we study translation invariance on the \hl{Tetrominoes} dataset \cite{multiobjectdatasets19} with simple geometric shapes over a black background. We compare Slot Attention \underline{(SA)} against Translation Invariant Slot Attention \underline{(\isat)}. We conduct experiments in two settings: (1) where only a few ($64$ to $1024$) training samples are available to the model, and (2) where training samples are biased to only have objects appear in the left side of the image and test samples may have objects in all positions. The former tests sample efficiency and the latter tests out-of-distribution generalization at test time.

Results are shown in Figure~\ref{fig:tetro}.
It can be seen that using pose-relative position encoding substantially improves sample efficiency. In experiment (1), \isat requires around $2\times$ to $4\times$ fewer samples to achieve performance equivalent to SA. Note that we used translational data augmentation in this experiment and picked optimal hyperparameters that favoured the SA baseline. This demonstrates that sample efficiency gains achieved through our method are not superseded by those from standard data augmentation. We argue that this may be due to how \isat models translation of \emph{individual objects}, which cannot be captured by simply translating the entire image.
In experiment (2), \isat again outperforms the SA baseline, demonstrating stronger out-of-distribution generalization. No data augmentation was used in this experiment in order to carefully control the distribution shift.

\subsection{Evaluating translation and scaling invariance}

We evaluate translation invariant SA, \underline{\isat}, and translation and scaling invariant SA, \underline{\isats}, on four synthetic multi-object datasets for unsupervised scene decomposition.

\begin{figure}[t!]
    \centering
    \includegraphics[width=0.9\linewidth]{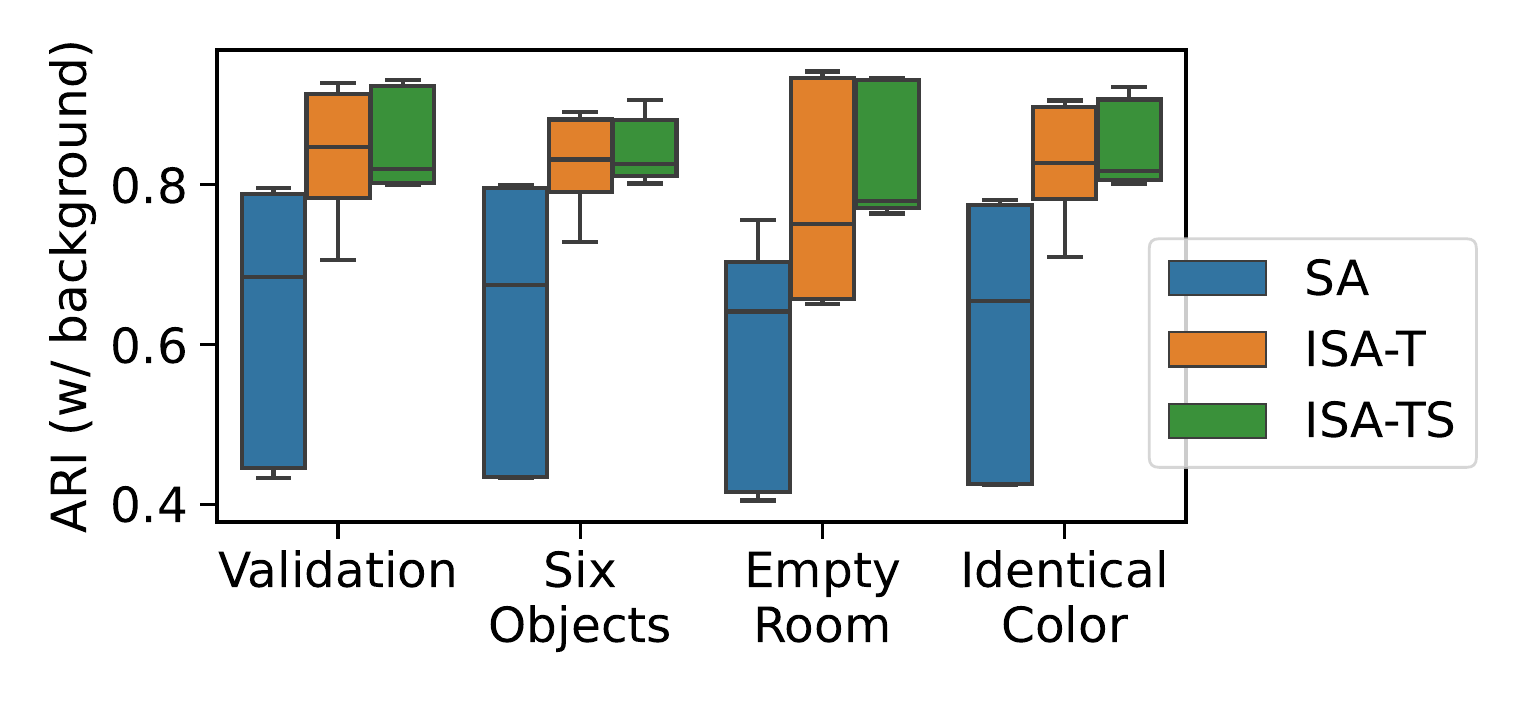}\\
    \includegraphics[width=0.9\linewidth]{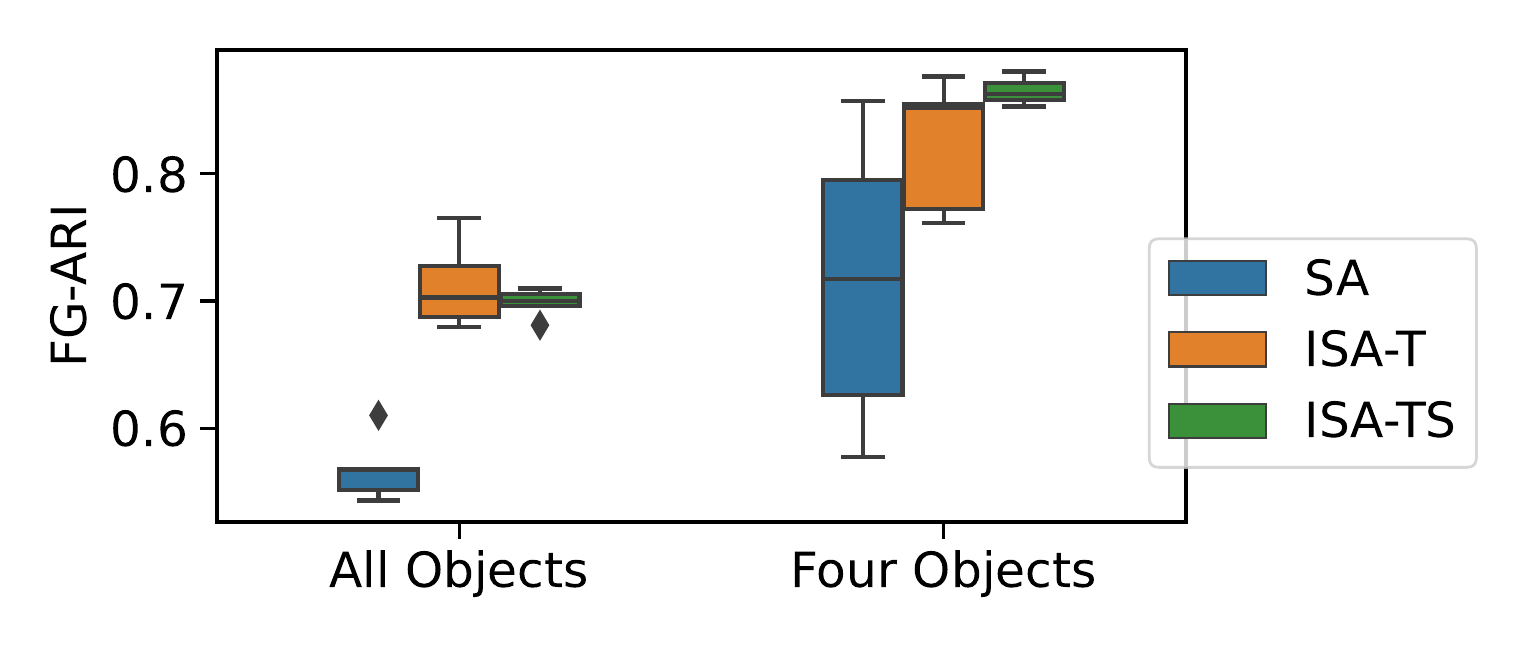}
    \vspace{-1em}
    
    \caption{\textbf{Top: Objects Room} ARI with background segments included: Higher is better. We use 10 seeds per experiment. For exact numbers see \cref{tab:objects_room}. \textbf{Bottom: MultiShapeNet} FG-ARI results using 5 random seeds. For exact numbers see \cref{tab:msne}.}
    \label{fig:msneplusobjsrm}
\end{figure}

In \hl{Objects Room} \cite{burgess19monet}, as shown in \cref{fig:msneplusobjsrm}, both invariant models achieve higher ARI scores across all validation sets when compared to SA. We treat floors, walls and ceilings as objects in these experiments and hence report ARI instead of FG-ARI. We also note a substantial reduction in uncertainity across seeds for ISA.

In \hl{MultiShapeNet} \cite{stelzner21decomposing}, we find both  \isat and \isats to greatly improve performance compared to SA (\cref{fig:msneplusobjsrm}-bottom, \emph{All Objects}). Upon closer qualitative examination, we found that \isats was more prone to over-segmentation; such as segmenting the head of a chair and its legs as separate objects. We controlled for this behavior by creating a new training and evaluation split consisting of all MultiShapeNet images with exactly four objects (\cref{fig:msneplusobjsrm}-bottom, \emph{Four Objects}). Here we set the number of slots to five making over-segmentation less likely. In this controlled setting, ISA-TS outperforms ISA-T and SA.

We also evaluate on the \hl{CLEVR} dataset \cite{johnson17clevr,multiobjectdatasets19}, but both ISA variants and the baseline Slot Attention model achieve near-perfect segmentation and decoding, see appendix table~\ref{tab:clevr}.

\begin{table}[t!]
    \centering
    \caption{CLEVRTex FG-ARI(\%) results on the test set, CAMO set (objects and backgrounds blend together) and OOD set (novel textures). Prior results taken from \citep{karazija21clevrtex} use 3 random seeds, we use 10 random seeds. FG-ARI is reported in \%. For MSE please see the \cref{tab:clevrtex:full}. (CNN) refers to models using a 4-layer CNN backbone, while (ResNet) models use a ResNet-34.}
    \vspace{2mm}
    \begin{tabular}{llll}
        \toprule
        \textbf{Method} & \textbf{Main} & \textbf{CAMO} & \textbf{OOD} \\
        \midrule
        \begin{filecontents*}{csv/clevrtex_baselines_2_short.tex}
method,test_miou,test_miou_sd,test_fg_ari,test_fg_ari_sd,test_mse,test_mse_sd,camo_miou,camo_miou_sd,camo_fg_ari,camo_fg_ari_sd,camo_mse,camo_mse_sd,ood_miou,ood_miou_sd,ood_fg_ari,ood_fg_ari_sd,ood_mse,ood_mse_sd
SPACE,9.1,3.5,17.5,4.1,298,80,8.7,3.5,10.6,2.1,251,61,6.9,3.3,12.7,3.4,387,66
DTI,33.8,1.3,79.9,1.4,438,22,27.5,1.6,72.9,1.9,377,17,32.6,1.1,73.7,1.0,590,4
AST-Seg-B3-CT,79.6,0.5,94.8,0.5,139,7,73.1,0.7,87.3,3.8,145,6,67.5,0.8,83.1,0.8,832,24
\end{filecontents*}
\csvreader[
            column count=11,
            head to column names,
            late after line=\\
        ]{csv/clevrtex_baselines_2_short.tex}{1=\one, 4=\two, 5=\three, 10=\six, 11=\seven, 16=\ten, 17=\eleven}
        {\one & \ebar{\two}{\three} & \ebar{\six}{\seven} & \ebar{\ten}{\eleven}}
        \midrule
        \begin{filecontents*}{csv/clevrtex_simplecnn_2_short.tex}
method,test_miou,test_miou_sd,test_fg_ari,test_fg_ari_sd,test_mse,test_mse_sd,camo_miou,camo_miou_sd,camo_fg_ari,camo_fg_ari_sd,camo_mse,camo_mse_sd,ood_miou,ood_miou_sd,ood_fg_ari,ood_fg_ari_sd,ood_mse,ood_mse_sd
SA (CNN),-,-,54.5,1.6,241.2,14.0,-,-,53.0,1.6,216.6,11.8,-,-,54.2,2.6,\underline{414.7},27.7
\isat (CNN),-,-,66.8,5.7,229.8,20.4,-,-,65.0,4.9,213.2,16.3,-,-,65.1,4.8,458.6,25.2
\isats (CNN),-,-,78.8,3.9,223.8,6.5,-,-,72.9,3.5,210.9,6.2,-,-,73.2,3.1,480.6,32.6
\end{filecontents*}
\csvreader[
            column count=11,
            head to column names,
            late after line=\\
        ]{csv/clevrtex_simplecnn_2_short.tex}{1=\one, 4=\two, 5=\three, 10=\six, 11=\seven, 16=\ten, 17=\eleven}
        {\one & \ebar{\two}{\three} & \ebar{\six}{\seven} & \ebar{\ten}{\eleven}}
        \midrule
        \begin{filecontents*}{csv/clevrtex_resnet_2_short.tex}
method,test_miou,test_miou_sd,test_fg_ari,test_fg_ari_sd,test_mse,test_mse_sd,camo_miou,camo_miou_sd,camo_fg_ari,camo_fg_ari_sd,camo_mse,camo_mse_sd,ood_miou,ood_miou_sd,ood_fg_ari,ood_fg_ari_sd,ood_mse,ood_mse_sd
SA (ResNet),-,-,91.3,2.7,206.7,20.8,-,-,84.9,2.9,224.4,21.9,-,-,81.4,1.4,629.1,29.9
\isat (ResNet),-,-,87.4,6.6,197.1,27.0,-,-,79.0,5.9,217.6,24.5,-,-,78.6,4.9,548.1,21.3
\isats (ResNet),-,-,92.9,0.4,176.9,3.4,-,-,86.2,0.8,196.3,7.1,-,-,84.4,0.8,578.3,22.5
\end{filecontents*}
\csvreader[
            column count=11,
            head to column names,
            late after line=\\
        ]{csv/clevrtex_resnet_2_short.tex}{1=\one, 4=\two, 5=\three, 10=\six, 11=\seven, 16=\ten, 17=\eleven}
        {\one & \ebar{\two}{\three} & \ebar{\six}{\seven} & \ebar{\ten}{\eleven}}
        \bottomrule
    \end{tabular}
    \label{tab:clevrtex:short}
\end{table}

\hl{CLEVRTex} \cite{karazija21clevrtex} is perhaps the most complex synthetic dataset in the relevant literature, where plain Slot Attention was previously reported to fail; we report the results of 12 different Invariant Slot Attention variants and ablations in the appendix in Table \ref{tab:clevrtex:full} and summarize key findings in Table \ref{tab:clevrtex:short}. Following the trend on previous datasets, \isat and \isats each lead to 10\%+ absolute improvement in FG-ARI (Table \ref{tab:clevrtex:short}, CNN).

To our surprise, plain Slot Attention can be made a competitive baseline with three simple changes: a stronger backbone (ResNet-34 \textit{without pre-training})~\citep{he2016deep}, a larger feature map resolution ($16^2$) with the same resolution in the encoder output and decoder input (i.e. the spatial broadcast resolution in the decoder is set to $16^2$ as well), and learnable initial slots (please see Section \ref{appendix:model:arch}).

Even in this improved setting (Table \ref{tab:clevrtex:short}, ResNet), we measure between 1\% and 3\% FG-ARI improvement between SA and \isats depending on the validation set. \isats outperforms the state-of-the-art (AST-Seg-B3-CT) on the out-of-distribution (OOD) test set, where \textit{novel textures and shapes} are used~\citep{sauvalle22unsupervised}. Our results suggest that ImageNet and background model pre-training, which are used in AST-Seg-B3-CT, are not necessary to solve CLEVRTex.
\isat, however, performs poorly in this setting. We note how its FG-ARI is being pulled down by a poorly performing seed, although that alone did not explain the overall drop.

In summary, we find that both translation and scaling symmetries can lead to large benefits in Slot Attention's ability to discover and segment objects. We hypothesize that this is in part because of the added inductive bias that guides the unsupervised discovery of objects, and partly due to weight sharing across positions and scales when decoding objects.

\subsection{Invariance to rotations}
\begin{figure}[t!]
    \centering
    \includegraphics[width=\linewidth]{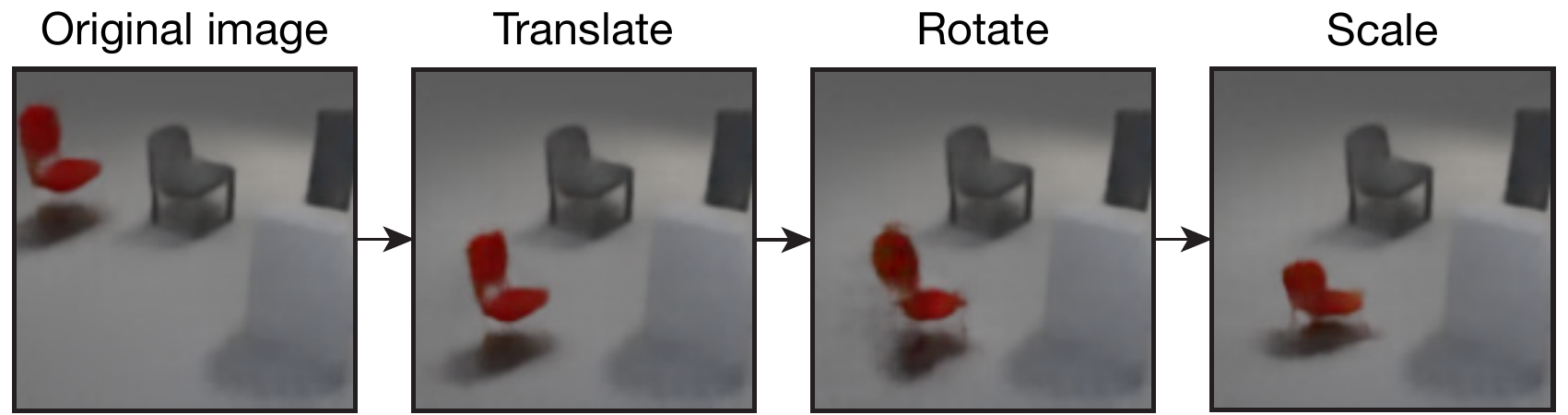}
    \vspace{-1em}
    \caption{Controlling slots with \isatsr on MultiShapeNet.}
    \label{fig:msne_traverse}
\end{figure}

\begin{table}[t]
\centering
\caption{\textbf{Rotation invariance}: Comparing \isats against \isatsr in various benchmarks. Objects Room results are ARIs whereas all others are FG-ARIs. Remaining benchmarks evaluations are in the appendix \cref{app:tab:rot}.}
\vspace{2mm}
\begin{tabular}{lllll}
    \toprule
                              & \multicolumn{2}{c}{(FG-)ARI $\uparrow$}  \\
    \textbf{Dataset}          & \isats  & \isatsr \\
    \midrule
    Objects Room (w/ bg) Val. & \ebar{85.5}{6.6} & \ebar{84.3}{4.6}\\
    CLEVR                     & \ebar{98.9}{0.2} & \ebar{98.0}{0.9} \\
    \midrule
    MultiShapeNet        &          &          \\
    - All Data                & \ebar{69.8}{1.1} & \ebar{77.7}{5.5} \\
    - Four Objects            & \ebar{86.5}{1.1} & \ebar{80.7}{6.4} \\
    \midrule
    CLEVRTex (CNN)    & \ebar{78.8}{3.9} & \ebar{79.6}{5.5} \\
    CLEVRTex (ResNet)        & \ebar{92.9}{0.4} & \ebar{93.3}{0.7} \\
    \bottomrule
\end{tabular}
\label{tab:rot_invariance}
\end{table}
We find rotational invariance \underline{\isatsr} leads to mixed results across datasets (see Table~\ref{tab:rot_invariance}. On one hand, \isatsr further pushes the state-of-the-art on CLEVRTex, increasing FG-ARI by around 0.5\%. On the other hand, we find that it decreases performance on Objects Room and CLEVR. Despite that, it does allow us to control the rotation of the decoded objects, which appears reasonably well captured (Figure \ref{fig:msne_traverse}). We hypothesize that our heuristic rotation estimator, that uses principal axes, is too susceptible to symmetric objects which cause ambiguity in the detected rotation. 
As a mitigation, future work could enable the model to predict corrections to the rotational heuristic.

\subsection{Real-world evaluation: Waymo Open}


\begin{figure}[t]
    \centering
    \includegraphics[width=1\linewidth]{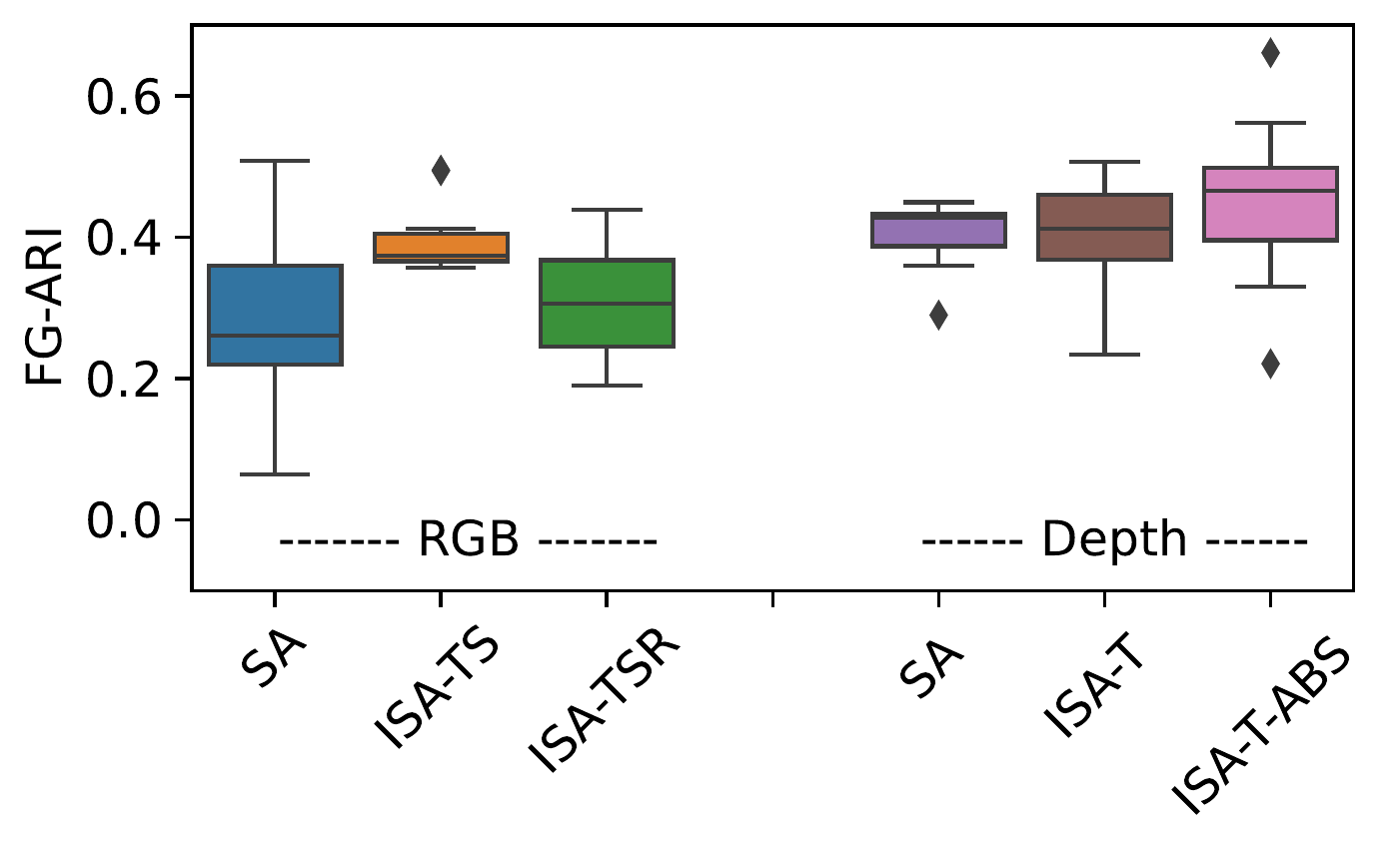}
    \vspace{-2em}
    \caption{\textbf{Waymo Open} (ResNet) across 10 seeds.} 
    \label{fig:waymo:plt}
\end{figure}

\begin{figure*}[h]
    \centering
    \includegraphics[width=1\linewidth]{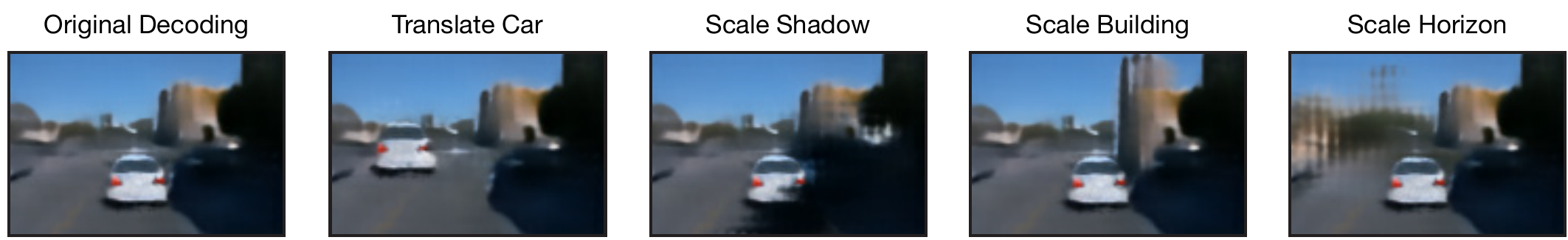}
    \vspace{-1em}
    \caption{Controlling slots with \isats (ResNet) on Waymo Open RGB.}
    \label{fig:waymo_traverse}
\end{figure*}

\begin{figure}[h!]
    \centering
    \includegraphics[width=0.7\linewidth]{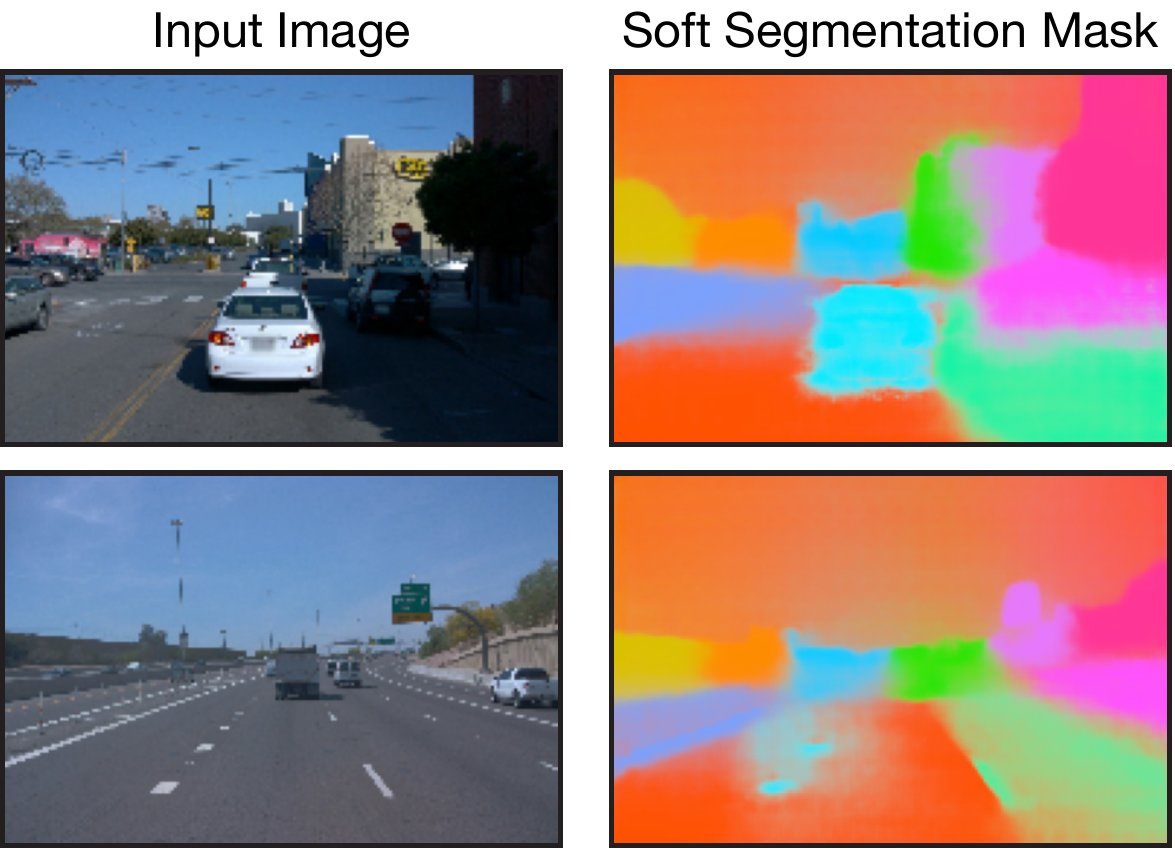}
    \vspace{-0.2em}
    \caption{Waymo Open RGB predicted soft segmentation masks of \isats (ResNet). Soft segmentation masks blend pixel colors based on normalized per-slot $\alpha$ masks.}
    \label{fig:waymo_segm}
\end{figure}


To investigate the utility of object pose invariance as an inductive bias on real-world visual data, we evaluate ISA for unsupervised instance segmentation on \hl{Waymo Open v1.4} \cite{sun20scalability,mei22waymo}, a dataset of videos collected from cameras on Waymo cars.
The dataset has been previously used in unsupervised learning (e.g.~\citet{elsayed22savipp}); however, we are the first to report positive single-frame RGB-only results for object discovery, i.e.~without the use of optical flow, depth features, or temporal information. Waymo Open is highly challenging since both the foreground objects and backgrounds are highly varied and move as the car moves. In comparison, prior methods tested on real-world datasets exploit static backgrounds \cite{kosiorek18sequential,sauvalle22unsupervised}.

\looseness=-1We find \isats to increase FG-ARI by 12\% compared to SA (Figure \ref{fig:waymo:plt}). Adding rotational symmetry decreases FG-ARI by about 5\% compared to \isats; we hypothesize that rotations of complex objects and backgrounds are too ambiguous to detect by our heuristic. 
Slot positions and scales in \isats allow controlling decoded outputs (Figure \ref{fig:waymo_traverse}).

Qualitatively, we find the computed FG-ARI and the perceived quality of predicted segmentation masks to not always be in agreement -- all Slot Attention models focus on landmarks, such as large buildings, trees, traffic signs and lanes of a road (Figure \ref{fig:waymo_segm}), whereas Waymo Open, naturally, only scores the detection of other cars and pedestrians.

Following \citet{elsayed22savipp}, we test if predicting depth instead of RGB improves emergent object segmentation. We find depth targets (the model still receives only RGB images as input) indeed increase the FG-ARI of Slot Attention, as cars and pedestrians do not blend in the clutter of complex backgrounds. But, depth images do not exhibit the same symmetries as RGB images. For example, as a car moves further away from the camera, its depth value changes whereas its color does not. Indeed, we find that \isat reaches the same performance as the baseline, unless we explicitly \textit{break the symmetry} in the model.

\subsection{Symmetry breaking}
Invariant Slot Attention ensures that the slot latent vectors are invariant to the absolute positions of the features they attend to\footnote{This holds true only with random initial slot statistics, since learnable initial slot statistics already break the symmetry.}, as long as the individual feature vectors do not already have position information encoded in them. However, we find that the visual backbone can leak absolute position information, which they can detect, for example by looking at the zero-padded edges added in their convolutions. This effect is limited to image boundary regions in the default 4-layer CNN backbone used in Slot Attention~\citep{locatello20object}, which we adopted for ISA, but it is clearly pronounced when using a more expressive ResNet-34 encoder.

Instead of Invariant Slot Attention having to ``cheat'' to break equivariance, we can explicitly append slot positions/rotations/scales to the slot latent vectors right before they are passed to the gated recurrent unit (GRU)~\citep{cho2014properties} in each Slot Attention iteration. The intuition for this design decision is that we allow the GRU to model the movement of per-slot attention masks over the iterations. We test this version of the model on CLEVRTex and find it to significantly improve ISA when using the default, shallow CNN backbone. We denote this model version using ``-Append''. Specifically, \isatsr-Append (85.4\% FG-ARI) outperforms \isatsr (79.6\% FG-ARI) and successfully segments various CLEVRTex scenes (Figure \ref{fig:clevrtex_segm}). When using a ResNet-34 backbone however, there is no significant difference between the two model variants, likely since the ResNet itself already leaks absolute position information into the model.

On Waymo Open, we further investigate breaking the symmetry in both encoder as well as the decoder by using both pose-relative as well as absolute position encoding. This leads to around 5\% FG-ARI improvement over both SA and T-SA (ResNet), Table \ref{tab:waymo}.

\begin{figure}[t!]
    \centering
    \includegraphics[width=0.8\linewidth]{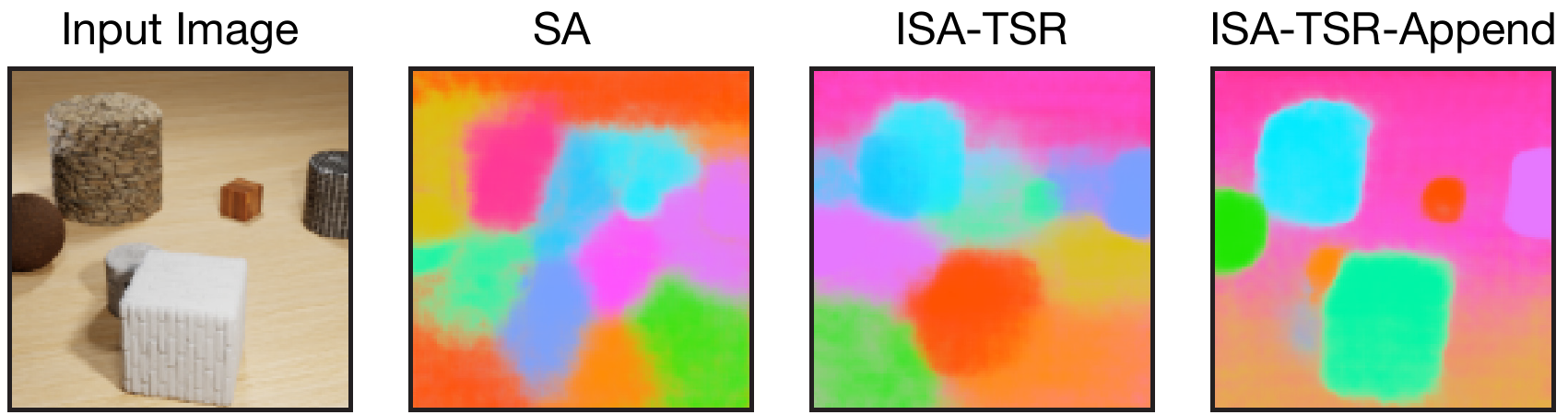}
    \vspace{-0.2em}
    \caption{CLEVRTex predicted segmentation masks of three CNN Slot Attention variants. The absolute difference in FG-ARI between the leftmost and rightmost model is 30\%.}
    \label{fig:clevrtex_segm}
\end{figure}

\subsection{Ablations}

\begin{table}[t]
    \centering
    \caption{\textbf{Model ablations}: We report FG-ARI (in \%) across Tetrominoes, CLEVRTex and Waymo Open for three different ISA model variants and two ablations: ``dec. only`` uses pose-relative position encoding only in the decoder and ``stop grad.`` does not backpropagate gradients through the estimated pose parameters.}
    \vspace{2mm}
    \begin{tabular}{lll}

    \toprule
    \textbf{Model} & \textbf{Dataset} & \textbf{FG-ARI}    \\
    \midrule
    \isat & \multirow{2}{*}{Tetrominoes}      & \ebar{96.3}{3.6} \\
    - dec. only  &  & \ebar{92.3}{2.4}   \\
    \midrule
    \isatsr (ResNet) & \multirow{3}{*}{CLEVRTex}       & \ebar{93.3}{0.7} \\
    - dec. only  &  & \ebar{93.4}{1.0}   \\
    - stop grad.  &  & \ebar{73.4}{10.5} \\
    \midrule
    \isats (ResNet) & \multirow{2}{*}{Waymo Open}        & \ebar{39.8}{5.3} \\
    - dec. only  & & \ebar{32.4}{7.1}   \\
    \bottomrule
    

    \end{tabular}
    \label{tab:ablations}
\end{table}

We conduct two ablation studies to examine whether spatial symmetries are necessary in the Slot Attention encoder module (``dec.~only'') and whether gradients need to flow through the computation of reference frame parameters $\left(\slotpositions, \slotscales, S_r\right)$ (``stop grad.''). Ablation results are summarized in Table~\ref{tab:ablations}.

``Dec.~only'' refers to using $\mathrm{abs\_grid}$ instead of $\mathrm{rel\_grid}$ in the attention mechanism but still using $\mathrm{rel\_grid}$ in the SB decoder. That is, the encoder is now not explicitly incorporating spatial symmetries whereas the decoder is still doing so. The slot reference frames required for doing so in the decoder are computed using the attention maps, $\mathrm{attn}$, from the last iteration of slot attention. We find that this ablation has a significant negative impact on segmentation performance on Tetrominoes, in which objects are symmetric under translation within image boundaries (excl.~occlusion). 
On CLEVRTex, we find its FG-ARI to be comparable to the unablated version (\isatsr). We find, however, that this ablation can have a negative effect on reconstruction quality: whereas ISA-TSR (ResNet) achieves a reconstruction MSE of $185.9\pm6.4$ on CLEVTex (Main), the  ``Dec.~only'' ablation results in an increased error of $200.4\pm6.6$. Since ``dec.~only'' does not decompose appearance and pose during encoding, we hypothesize that this entangled slot representation is detrimental to generalizing across object positions and textures. Similar to Tetrominoes, we find a significant reduction in FG-ARI for this ablation on Waymo Open.

The ``stop grad.'' results in Table \ref{tab:ablations}, show that is it crucial to allow the gradients to flow through the pose estimation mechanism, which suggest that the backwards pathway alters cross-attention weights in Slot Attention to obtain more useful reference frames.

\section{Conclusion}
\label{sec:conclusion}
We have introduced Invariant Slot Attention (ISA), a method for unsupervised scene decomposition and object discovery using slot-centric reference frames, which enables learning object representations that are invariant to geometric transformation including translation, scale and rotation.
Our method incorporates spatial symmetries on a per-object basis with little computational overhead, which is achieved via simple changes to the positional encoding used both in the attention mechanism and the decoder of Slot Attention. 

\looseness=-1Even though our investigation of data efficiency and generalization benefits focused on unsupervised scene understanding, \textit{per-object} reference frames are likely worth broader consideration in future work, as data augmentation alone is unable to capture these symmetries effectively. One limitation of such an approach is that \textit{exact} translation, scale and rotation symmetries of objects are rarely present in images of 3D scenes, and it is unclear how backgrounds should be handled. Despite this, we see evidence that per-object symmetries, especially in combination with ways for the model to \textit{break the symmetry} by additionally using absolute, global reference frames, frequently improve model performance.

\section*{Acknowledgements}

We would like to thank Mike Mozer for helpful discussions and feedback on the manuscript, and Klaus Greff for assistance with Kubric-generated datasets \citep{greff22kubric}. We would also like to thank the anonymous reviewers of the NeurIPS'22 NeurReps workshop and ICML'23.

\newpage

\bibliography{main,position_encoding}

\begin{thebibliography}{81}
\providecommand{\natexlab}[1]{#1}
\providecommand{\url}[1]{\texttt{#1}}
\expandafter\ifx\csname urlstyle\endcsname\relax
  \providecommand{\doi}[1]{doi: #1}\else
  \providecommand{\doi}{doi: \begingroup \urlstyle{rm}\Url}\fi

\bibitem[Barlow(2009)]{barlow09grandmother}
Barlow, H.
\newblock Grandmother cells, symmetry, and invariance: How the term arose and
  what the facts suggest.
\newblock In Gazzaniga, M.~S. (ed.), \emph{The Cognitive Neurosciences}, pp.\
  309--320. The MIT Press, fourth edition, 2009.

\bibitem[Bello et~al.(2019)Bello, Zoph, Le, Vaswani, and
  Shlens]{bello19attention}
Bello, I., Zoph, B., Le, Q., Vaswani, A., and Shlens, J.
\newblock Attention {{Augmented Convolutional Networks}}.
\newblock In \emph{ICCV}, 2019.

\bibitem[Bottini \& Doeller(2020)Bottini and Doeller]{bottini2020knowledge}
Bottini, R. and Doeller, C.~F.
\newblock Knowledge across reference frames: Cognitive maps and image spaces.
\newblock \emph{Trends in Cognitive Sciences}, 24\penalty0 (8):\penalty0
  606--619, 2020.

\bibitem[Bronstein et~al.(2021)Bronstein, Bruna, Cohen, and
  Velickovic]{bronstein21geometric}
Bronstein, M.~M., Bruna, J., Cohen, T., and Velickovic, P.
\newblock Geometric deep learning: Grids, groups, graphs, geodesics, and
  gauges.
\newblock \emph{CoRR}, abs/2104.13478, 2021.

\bibitem[Burgess et~al.(2019)Burgess, Matthey, Watters, Kabra, Higgins,
  Botvinick, and Lerchner]{burgess19monet}
Burgess, C.~P., Matthey, L., Watters, N., Kabra, R., Higgins, I., Botvinick,
  M.~M., and Lerchner, A.
\newblock Monet: Unsupervised scene decomposition and representation.
\newblock \emph{CoRR}, abs/1901.11390, 2019.

\bibitem[Carion et~al.(2020)Carion, Massa, Synnaeve, Usunier, Kirillov, and
  Zagoruyko]{carion20endtoend}
Carion, N., Massa, F., Synnaeve, G., Usunier, N., Kirillov, A., and Zagoruyko,
  S.
\newblock End-to-{{End Object Detection}} with {{Transformers}}.
\newblock In \emph{ECCV}, 2020.

\bibitem[Cho et~al.(2014)Cho, Van~Merri{\"e}nboer, Bahdanau, and
  Bengio]{cho2014properties}
Cho, K., Van~Merri{\"e}nboer, B., Bahdanau, D., and Bengio, Y.
\newblock On the properties of neural machine translation: Encoder-decoder
  approaches.
\newblock \emph{arXiv preprint arXiv:1409.1259}, 2014.

\bibitem[Cordonnier et~al.(2020)Cordonnier, Loukas, and
  Jaggi]{cordonnier2020relationship}
Cordonnier, J.-B., Loukas, A., and Jaggi, M.
\newblock On the relationship between self-attention and convolutional layers.
\newblock In \emph{ICLR}, 2020.

\bibitem[Crawford \& Pineau(2019)Crawford and Pineau]{crawford19spatially}
Crawford, E. and Pineau, J.
\newblock Spatially invariant unsupervised object detection with convolutional
  neural networks.
\newblock In \emph{AAAI}, 2019.

\bibitem[Dosovitskiy et~al.(2021)Dosovitskiy, Beyer, Kolesnikov, Weissenborn,
  Zhai, Unterthiner, Dehghani, Minderer, Heigold, Gelly, Uszkoreit, and
  Houlsby]{dosovitskiy21image}
Dosovitskiy, A., Beyer, L., Kolesnikov, A., Weissenborn, D., Zhai, X.,
  Unterthiner, T., Dehghani, M., Minderer, M., Heigold, G., Gelly, S.,
  Uszkoreit, J., and Houlsby, N.
\newblock An {{Image}} is {{Worth}} 16x16 {{Words}}: {{Transformers}} for
  {{Image Recognition}} at {{Scale}}.
\newblock In \emph{ICLR}, 2021.

\bibitem[Dufter et~al.(2022)Dufter, Schmitt, and Sch{\"u}tze]{dufter22position}
Dufter, P., Schmitt, M., and Sch{\"u}tze, H.
\newblock Position {{Information}} in {{Transformers}}: {{An Overview}}.
\newblock \emph{Comput. Linguistics}, 48\penalty0 (3):\penalty0 733--763, 2022.

\bibitem[Elsayed et~al.(2022)Elsayed, Mahendran, van Steenkiste, Greff, Mozer,
  and Kipf]{elsayed22savipp}
Elsayed, G.~F., Mahendran, A., van Steenkiste, S., Greff, K., Mozer, M.~C., and
  Kipf, T.
\newblock {SAVi++}: Towards end-to-end object-centric learning from real-world
  videos.
\newblock \emph{NeurIPS}, 2022.

\bibitem[Emami et~al.(2021)Emami, He, Ranka, and Rangarajan]{emami21efficient}
Emami, P., He, P., Ranka, S., and Rangarajan, A.
\newblock Efficient iterative amortized inference for learning symmetric and
  disentangled multi-object representations.
\newblock In \emph{ICML}, 2021.

\bibitem[Engelcke et~al.(2020)Engelcke, Kosiorek, Jones, and
  Posner]{engelcke20genesis}
Engelcke, M., Kosiorek, A.~R., Jones, O.~P., and Posner, I.
\newblock {GENESIS:} generative scene inference and sampling with
  object-centric latent representations.
\newblock In \emph{ICLR}, 2020.

\bibitem[Eslami et~al.(2016)Eslami, Heess, Weber, Tassa, Szepesvari,
  Kavukcuoglu, and Hinton]{eslami16attend}
Eslami, S. M.~A., Heess, N., Weber, T., Tassa, Y., Szepesvari, D., Kavukcuoglu,
  K., and Hinton, G.~E.
\newblock Attend, infer, repeat: Fast scene understanding with generative
  models.
\newblock In \emph{NeurIPS}, 2016.

\bibitem[Gao et~al.(2021)Gao, Zheng, Wang, Dai, and Li]{gao21fast}
Gao, P., Zheng, M., Wang, X., Dai, J., and Li, H.
\newblock Fast {{Convergence}} of {{DETR}} with {{Spatially Modulated
  Co-Attention}}.
\newblock In \emph{ICCV}, 2021.

\bibitem[Greff et~al.(2015)Greff, Srivastava, and Schmidhuber]{greff15binding}
Greff, K., Srivastava, R.~K., and Schmidhuber, J.
\newblock Binding via reconstruction clustering.
\newblock \emph{CoRR}, abs/1511.06418, 2015.

\bibitem[Greff et~al.(2016)Greff, Rasmus, Berglund, Hao, Valpola, and
  Schmidhuber]{greff16tagger}
Greff, K., Rasmus, A., Berglund, M., Hao, T.~H., Valpola, H., and Schmidhuber,
  J.
\newblock Tagger: Deep unsupervised perceptual grouping.
\newblock In Lee, D.~D., Sugiyama, M., von Luxburg, U., Guyon, I., and Garnett,
  R. (eds.), \emph{NeurIPS}, 2016.

\bibitem[Greff et~al.(2017)Greff, van Steenkiste, and
  Schmidhuber]{greff17neural}
Greff, K., van Steenkiste, S., and Schmidhuber, J.
\newblock Neural expectation maximization.
\newblock In Guyon, I., von Luxburg, U., Bengio, S., Wallach, H.~M., Fergus,
  R., Vishwanathan, S. V.~N., and Garnett, R. (eds.), \emph{NeurIPS}, 2017.

\bibitem[Greff et~al.(2019)Greff, Kaufman, Kabra, Watters, Burgess, Zoran,
  Matthey, Botvinick, and Lerchner]{greff19multiobject}
Greff, K., Kaufman, R.~L., Kabra, R., Watters, N., Burgess, C., Zoran, D.,
  Matthey, L., Botvinick, M.~M., and Lerchner, A.
\newblock Multi-object representation learning with iterative variational
  inference.
\newblock In \emph{ICML}, 2019.

\bibitem[Greff et~al.(2020)Greff, Van~Steenkiste, and
  Schmidhuber]{greff2020binding}
Greff, K., Van~Steenkiste, S., and Schmidhuber, J.
\newblock On the binding problem in artificial neural networks.
\newblock \emph{arXiv preprint arXiv:2012.05208}, 2020.

\bibitem[Greff et~al.(2022)Greff, Belletti, Beyer, Doersch, Du, Duckworth,
  Fleet, Gnanapragasam, Golemo, Herrmann, Kipf, Kundu, Lagun, Laradji, Liu,
  Meyer, Miao, Nowrouzezahrai, {\"{O}}ztireli, Pot, Radwan, Rebain, Sabour,
  Sajjadi, Sela, Sitzmann, Stone, Sun, Vora, Wang, Wu, Yi, Zhong, and
  Tagliasacchi]{greff22kubric}
Greff, K., Belletti, F., Beyer, L., Doersch, C., Du, Y., Duckworth, D., Fleet,
  D.~J., Gnanapragasam, D., Golemo, F., Herrmann, C., Kipf, T., Kundu, A.,
  Lagun, D., Laradji, I.~H., Liu, H.~D., Meyer, H., Miao, Y., Nowrouzezahrai,
  D., {\"{O}}ztireli, A.~C., Pot, E., Radwan, N., Rebain, D., Sabour, S.,
  Sajjadi, M. S.~M., Sela, M., Sitzmann, V., Stone, A., Sun, D., Vora, S.,
  Wang, Z., Wu, T., Yi, K.~M., Zhong, F., and Tagliasacchi, A.
\newblock Kubric: {A} scalable dataset generator.
\newblock In \emph{CVPR}, 2022.

\bibitem[Han et~al.(2022)Han, Rong, Xu, Sun, and Huang]{han22equivariant}
Han, J., Rong, Y., Xu, T., Sun, F., and Huang, W.
\newblock Equivariant graph hierarchy-based neural networks.
\newblock \emph{CoRR}, abs/2202.10643, 2022.

\bibitem[Hawkins et~al.(2019)Hawkins, Lewis, Klukas, Purdy, and
  Ahmad]{hawking19framework}
Hawkins, J., Lewis, M., Klukas, M., Purdy, S., and Ahmad, S.
\newblock A framework for intelligence and cortical function based on grid
  cells in the neocortex.
\newblock \emph{Frontiers in Neural Circuits}, 12, 2019.

\bibitem[He et~al.(2016)He, Zhang, Ren, and Sun]{he2016deep}
He, K., Zhang, X., Ren, S., and Sun, J.
\newblock Deep residual learning for image recognition.
\newblock In \emph{CVPR}, 2016.

\bibitem[Hinton(1981)]{hinton81parallel}
Hinton, G.~E.
\newblock A parallel computation that assigns canonical object-based frames of
  reference.
\newblock In Hayes, P.~J. (ed.), \emph{Proceedings of the 7th International
  Joint Conference on Artificial Intelligence, {IJCAI} '81}, 1981.

\bibitem[Hinton(2021)]{hinton21glom}
Hinton, G.~E.
\newblock How to represent part-whole hierarchies in a neural network.
\newblock \emph{CoRR}, abs/2102.12627, 2021.

\bibitem[Hinton et~al.(2011)Hinton, Krizhevsky, and Wang]{hinton11transforming}
Hinton, G.~E., Krizhevsky, A., and Wang, S.~D.
\newblock Transforming auto-encoders.
\newblock In Honkela, T., Duch, W., Girolami, M.~A., and Kaski, S. (eds.),
  \emph{International Conference on Artificial Neural Networks}, 2011.

\bibitem[Hinton et~al.(2018)Hinton, Sabour, and Frosst]{hinton18matrix}
Hinton, G.~E., Sabour, S., and Frosst, N.
\newblock Matrix capsules with {EM} routing.
\newblock In \emph{ICLR}, 2018.

\bibitem[Hubert \& Arabie(1985)Hubert and Arabie]{hubert1985comparing}
Hubert, L. and Arabie, P.
\newblock Comparing partitions.
\newblock \emph{Journal of Classification}, 1985.

\bibitem[Ioffe \& Szegedy(2015)Ioffe and Szegedy]{ioffe2015batch}
Ioffe, S. and Szegedy, C.
\newblock Batch normalization: Accelerating deep network training by reducing
  internal covariate shift.
\newblock In \emph{ICML}, 2015.

\bibitem[Jaderberg et~al.(2015)Jaderberg, Simonyan, Zisserman, and
  Kavukcuoglu]{jaderberg15spatial}
Jaderberg, M., Simonyan, K., Zisserman, A., and Kavukcuoglu, K.
\newblock Spatial transformer networks.
\newblock In \emph{NeurIPS}, 2015.

\bibitem[Jiang \& Ahn(2020)Jiang and Ahn]{jiang20generative}
Jiang, J. and Ahn, S.
\newblock Generative neurosymbolic machines.
\newblock In \emph{NeurIPS}, 2020.

\bibitem[Jiang et~al.(2020)Jiang, Janghorbani, de~Melo, and Ahn]{jiang20scalor}
Jiang, J., Janghorbani, S., de~Melo, G., and Ahn, S.
\newblock {SCALOR:} generative world models with scalable object
  representations.
\newblock In \emph{ICLR}, 2020.

\bibitem[Johnson et~al.(2017)Johnson, Hariharan, van~der Maaten, Fei{-}Fei,
  Zitnick, and Girshick]{johnson17clevr}
Johnson, J., Hariharan, B., van~der Maaten, L., Fei{-}Fei, L., Zitnick, C.~L.,
  and Girshick, R.~B.
\newblock {CLEVR:} {A} diagnostic dataset for compositional language and
  elementary visual reasoning.
\newblock In \emph{CVPR}, 2017.

\bibitem[Kabra et~al.(2019)Kabra, Burgess, Matthey, Kaufman, Greff, Reynolds,
  and Lerchner]{multiobjectdatasets19}
Kabra, R., Burgess, C., Matthey, L., Kaufman, R.~L., Greff, K., Reynolds, M.,
  and Lerchner, A.
\newblock Multi-object datasets.
\newblock https://github.com/deepmind/multi-object-datasets/, 2019.

\bibitem[Karazija et~al.(2021)Karazija, Laina, and
  Rupprecht]{karazija21clevrtex}
Karazija, L., Laina, I., and Rupprecht, C.
\newblock Clevrtex: {A} texture-rich benchmark for unsupervised multi-object
  segmentation.
\newblock In \emph{NeurIPS Track on Datasets and Benchmarks 1}, 2021.

\bibitem[Kingma \& Ba(2015)Kingma and Ba]{kingma15adam}
Kingma, D.~P. and Ba, J.
\newblock Adam: {A} method for stochastic optimization.
\newblock In Bengio, Y. and LeCun, Y. (eds.), \emph{ICLR}, 2015.

\bibitem[Kipf et~al.(2022)Kipf, Elsayed, Mahendran, Stone, Sabour, Heigold,
  Jonschkowski, Dosovitskiy, and Greff]{kipf22conditional}
Kipf, T., Elsayed, G.~F., Mahendran, A., Stone, A., Sabour, S., Heigold, G.,
  Jonschkowski, R., Dosovitskiy, A., and Greff, K.
\newblock Conditional object-centric learning from video.
\newblock In \emph{ICLR}, 2022.

\bibitem[Kosiorek et~al.(2018)Kosiorek, Kim, Teh, and
  Posner]{kosiorek18sequential}
Kosiorek, A.~R., Kim, H., Teh, Y.~W., and Posner, I.
\newblock Sequential attend, infer, repeat: Generative modelling of moving
  objects.
\newblock In \emph{NeurIPS}, 2018.

\bibitem[Kosiorek et~al.(2019)Kosiorek, Sabour, Teh, and
  Hinton]{kosiorek19stacked}
Kosiorek, A.~R., Sabour, S., Teh, Y.~W., and Hinton, G.~E.
\newblock Stacked capsule autoencoders.
\newblock In Wallach, H.~M., Larochelle, H., Beygelzimer, A.,
  d'Alch{\'{e}}{-}Buc, F., Fox, E.~B., and Garnett, R. (eds.), \emph{NeurIPS},
  pp.\  15486--15496, 2019.

\bibitem[Lin et~al.(2020)Lin, Wu, Peri, Sun, Singh, Deng, Jiang, and
  Ahn]{lin20space}
Lin, Z., Wu, Y., Peri, S.~V., Sun, W., Singh, G., Deng, F., Jiang, J., and Ahn,
  S.
\newblock {SPACE:} unsupervised object-oriented scene representation via
  spatial attention and decomposition.
\newblock In \emph{ICLR}, 2020.

\bibitem[Liu et~al.(2022)Liu, Li, Zhang, Yang, Qi, Su, Zhu, and
  Zhang]{liu22dabdetr}
Liu, S., Li, F., Zhang, H., Yang, X., Qi, X., Su, H., Zhu, J., and Zhang, L.
\newblock {{DAB-DETR}}: {{Dynamic Anchor Boxes}} are {{Better Queries}} for
  {{DETR}}.
\newblock In \emph{ICLR}, 2022.

\bibitem[Locatello et~al.(2020)Locatello, Weissenborn, Unterthiner, Mahendran,
  Heigold, Uszkoreit, Dosovitskiy, and Kipf]{locatello20object}
Locatello, F., Weissenborn, D., Unterthiner, T., Mahendran, A., Heigold, G.,
  Uszkoreit, J., Dosovitskiy, A., and Kipf, T.
\newblock Object-centric learning with slot attention.
\newblock In \emph{NeurIPS}, 2020.

\bibitem[Loshchilov \& Hutter(2017)Loshchilov and Hutter]{loshchilov17sgdr}
Loshchilov, I. and Hutter, F.
\newblock {SGDR:} stochastic gradient descent with warm restarts.
\newblock In \emph{ICLR}, 2017.

\bibitem[Luong et~al.(2015)Luong, Pham, and Manning]{luong15effective}
Luong, T., Pham, H., and Manning, C.~D.
\newblock Effective approaches to attention-based neural machine translation.
\newblock In \emph{EMNLP}, 2015.

\bibitem[Mei et~al.(2022)Mei, Zhu, Yan, Yan, Qiao, Chen, and
  Kretzschmar]{mei22waymo}
Mei, J., Zhu, A.~Z., Yan, X., Yan, H., Qiao, S., Chen, L., and Kretzschmar, H.
\newblock Waymo open dataset: Panoramic video panoptic segmentation.
\newblock In \emph{ECCV}, 2022.

\bibitem[Meng et~al.(2021)Meng, Chen, Fan, Zeng, Li, Yuan, Sun, and
  Wang]{meng21conditional}
Meng, D., Chen, X., Fan, Z., Zeng, G., Li, H., Yuan, Y., Sun, L., and Wang, J.
\newblock Conditional {{DETR}} for {{Fast Training Convergence}}.
\newblock In \emph{ICCV}, 2021.

\bibitem[Mildenhall et~al.(2020)Mildenhall, Srinivasan, Tancik, Barron,
  Ramamoorthi, and Ng]{mildenhall20nerf}
Mildenhall, B., Srinivasan, P.~P., Tancik, M., Barron, J.~T., Ramamoorthi, R.,
  and Ng, R.
\newblock Nerf: Representing scenes as neural radiance fields for view
  synthesis.
\newblock In \emph{ECCV}, 2020.

\bibitem[Monnier et~al.(2021)Monnier, Vincent, Ponce, and
  Aubry]{monnier21unsupervised}
Monnier, T., Vincent, E., Ponce, J., and Aubry, M.
\newblock Unsupervised layered image decomposition into object prototypes.
\newblock In \emph{ICCV}, 2021.

\bibitem[Nair \& Hinton(2010)Nair and Hinton]{nair2010rectified}
Nair, V. and Hinton, G.~E.
\newblock Rectified linear units improve restricted boltzmann machines.
\newblock In \emph{ICML}, 2010.

\bibitem[Park et~al.(2022)Park, Biza, Zhao, van~de Meent, and
  Walters]{park22learning}
Park, J.~Y., Biza, O., Zhao, L., van~de Meent, J., and Walters, R.
\newblock Learning symmetric embeddings for equivariant world models.
\newblock In \emph{ICML}, 2022.

\bibitem[Parmar et~al.(2019)Parmar, Ramachandran, Vaswani, Bello, Levskaya, and
  Shlens]{parmar19standalone}
Parmar, N., Ramachandran, P., Vaswani, A., Bello, I., Levskaya, A., and Shlens,
  J.
\newblock Stand-{{Alone Self-Attention}} in {{Vision Models}}.
\newblock In \emph{NeurIPS}, 2019.

\bibitem[Ranftl et~al.(2021)Ranftl, Bochkovskiy, and Koltun]{ranftl21vision}
Ranftl, R., Bochkovskiy, A., and Koltun, V.
\newblock Vision transformers for dense prediction.
\newblock In \emph{ICCV}, 2021.

\bibitem[Sabour et~al.(2017)Sabour, Frosst, and Hinton]{sabour17dynamic}
Sabour, S., Frosst, N., and Hinton, G.~E.
\newblock Dynamic routing between capsules.
\newblock In Guyon, I., von Luxburg, U., Bengio, S., Wallach, H.~M., Fergus,
  R., Vishwanathan, S. V.~N., and Garnett, R. (eds.), \emph{NeurIPS}, 2017.

\bibitem[Sajjadi et~al.(2021)Sajjadi, Meyer, Pot, Bergmann, Greff, Radwan,
  Vora, Lucic, Duckworth, Dosovitskiy, Uszkoreit, Funkhouser, and
  Tagliasacchi]{sajjadi21scene}
Sajjadi, M. S.~M., Meyer, H., Pot, E., Bergmann, U., Greff, K., Radwan, N.,
  Vora, S., Lucic, M., Duckworth, D., Dosovitskiy, A., Uszkoreit, J.,
  Funkhouser, T.~A., and Tagliasacchi, A.
\newblock Scene representation transformer: Geometry-free novel view synthesis
  through set-latent scene representations.
\newblock \emph{CoRR}, abs/2111.13152, 2021.

\bibitem[Sajjadi et~al.(2022)Sajjadi, Duckworth, Mahendran, van Steenkiste,
  Pavetic, Lucic, Guibas, Greff, and Kipf]{sajjadi22object}
Sajjadi, M. S.~M., Duckworth, D., Mahendran, A., van Steenkiste, S., Pavetic,
  F., Lucic, M., Guibas, L.~J., Greff, K., and Kipf, T.
\newblock Object scene representation transformer.
\newblock \emph{CoRR}, abs/2206.06922, 2022.

\bibitem[Sauvalle \& de~La~Fortelle(2022)Sauvalle and
  de~La~Fortelle]{sauvalle22unsupervised}
Sauvalle, B. and de~La~Fortelle, A.
\newblock Unsupervised multi-object segmentation using attention and
  soft-argmax.
\newblock \emph{CoRR}, abs/2205.13271, 2022.

\bibitem[Seitzer et~al.(2023)Seitzer, Horn, Zadaianchuk, Zietlow, Xiao,
  Simon-Gabriel, He, Zhang, Schölkopf, Brox, and Locatello]{seitzer23bridging}
Seitzer, M., Horn, M., Zadaianchuk, A., Zietlow, D., Xiao, T., Simon-Gabriel,
  C.-J., He, T., Zhang, Z., Schölkopf, B., Brox, T., and Locatello, F.
\newblock Bridging the gap to real-world object-centric learning.
\newblock In \emph{ICLR}, 2023.

\bibitem[Shaw et~al.(2018)Shaw, Uszkoreit, and Vaswani]{shaw18selfattention}
Shaw, P., Uszkoreit, J., and Vaswani, A.
\newblock Self-{{Attention}} with {{Relative Position Representations}}.
\newblock In \emph{NAACL-HLT}, 2018.

\bibitem[Singh et~al.(2021)Singh, Deng, and Ahn]{singh21illiterate}
Singh, G., Deng, F., and Ahn, S.
\newblock Illiterate {DALL-E} learns to compose.
\newblock \emph{CoRR}, abs/2110.11405, 2021.

\bibitem[Singh et~al.(2022)Singh, Wu, and Ahn]{singh22simple}
Singh, G., Wu, Y., and Ahn, S.
\newblock Simple unsupervised object-centric learning for complex and
  naturalistic videos.
\newblock \emph{CoRR}, abs/2205.14065, 2022.

\bibitem[Smirnov et~al.(2021)Smirnov, Gharbi, Fisher, Guizilini, Efros, and
  Solomon]{smirnov21marionette}
Smirnov, D., Gharbi, M., Fisher, M., Guizilini, V., Efros, A.~A., and Solomon,
  J.~M.
\newblock Marionette: Self-supervised sprite learning.
\newblock In \emph{NeurIPS}, 2021.

\bibitem[Srinivas et~al.(2021)Srinivas, Lin, Parmar, Shlens, Abbeel, and
  Vaswani]{srinivas21bottleneck}
Srinivas, A., Lin, T.-Y., Parmar, N., Shlens, J., Abbeel, P., and Vaswani, A.
\newblock Bottleneck {{Transformers}} for {{Visual Recognition}}.
\newblock In \emph{CVPR}, 2021.

\bibitem[Stelzner et~al.(2021)Stelzner, Kersting, and
  Kosiorek]{stelzner21decomposing}
Stelzner, K., Kersting, K., and Kosiorek, A.~R.
\newblock Decomposing 3d scenes into objects via unsupervised volume
  segmentation.
\newblock \emph{CoRR}, abs/2104.01148, 2021.

\bibitem[Sun et~al.(2020)Sun, Kretzschmar, Dotiwalla, Chouard, Patnaik, Tsui,
  Guo, Zhou, Chai, Caine, Vasudevan, Han, Ngiam, Zhao, Timofeev, Ettinger,
  Krivokon, Gao, Joshi, Zhang, Shlens, Chen, and Anguelov]{sun20scalability}
Sun, P., Kretzschmar, H., Dotiwalla, X., Chouard, A., Patnaik, V., Tsui, P.,
  Guo, J., Zhou, Y., Chai, Y., Caine, B., Vasudevan, V., Han, W., Ngiam, J.,
  Zhao, H., Timofeev, A., Ettinger, S., Krivokon, M., Gao, A., Joshi, A.,
  Zhang, Y., Shlens, J., Chen, Z., and Anguelov, D.
\newblock Scalability in perception for autonomous driving: Waymo open dataset.
\newblock In \emph{CVPR}, 2020.

\bibitem[van Steenkiste et~al.(2018)van Steenkiste, Chang, Greff, and
  Schmidhuber]{steenkiste18relational}
van Steenkiste, S., Chang, M., Greff, K., and Schmidhuber, J.
\newblock Relational neural expectation maximization: Unsupervised discovery of
  objects and their interactions.
\newblock In \emph{ICLR}, 2018.

\bibitem[Vaswani et~al.(2017)Vaswani, Shazeer, Parmar, Uszkoreit, Jones, Gomez,
  Kaiser, and Polosukhin]{vaswani17attention}
Vaswani, A., Shazeer, N., Parmar, N., Uszkoreit, J., Jones, L., Gomez, A.~N.,
  Kaiser, L., and Polosukhin, I.
\newblock Attention is {{All}} you {{Need}}.
\newblock In \emph{NeurIPS}, 2017.

\bibitem[Wang et~al.(2020{\natexlab{a}})Wang, Kohler, and Jr.]{wang20policy}
Wang, D., Kohler, C., and Jr., R.~P.
\newblock Policy learning in {SE(3)} action spaces.
\newblock In \emph{Robot Learning, CoRL}, 2020{\natexlab{a}}.

\bibitem[Wang et~al.(2020{\natexlab{b}})Wang, Zhu, Green, Adam, Yuille, and
  Chen]{wang20axialdeeplab}
Wang, H., Zhu, Y., Green, B., Adam, H., Yuille, A.~L., and Chen, L.-C.
\newblock Axial-{{DeepLab}}: {{Stand-Alone Axial-Attention}} for {{Panoptic
  Segmentation}}.
\newblock In \emph{ECCV}, 2020{\natexlab{b}}.

\bibitem[Wang et~al.(2022)Wang, Zhang, Yang, and Sun]{wang22anchor}
Wang, Y., Zhang, X., Yang, T., and Sun, J.
\newblock Anchor {{DETR}}: {{Query Design}} for {{Transformer-Based Detector}}.
\newblock In \emph{AAAI}, 2022.

\bibitem[Watters et~al.(2019)Watters, Matthey, Burgess, and
  Lerchner]{watters19spatial}
Watters, N., Matthey, L., Burgess, C.~P., and Lerchner, A.
\newblock Spatial broadcast decoder: {A} simple architecture for learning
  disentangled representations in vaes.
\newblock \emph{CoRR}, abs/1901.07017, 2019.

\bibitem[Wu \& He(2018)Wu and He]{wu2018group}
Wu, Y. and He, K.
\newblock Group normalization.
\newblock In \emph{ECCV}, 2018.

\bibitem[Wu et~al.(2022)Wu, Dvornik, Greff, Kipf, and Garg]{wu22slotformer}
Wu, Z., Dvornik, N., Greff, K., Kipf, T., and Garg, A.
\newblock Slotformer: Unsupervised visual dynamics simulation with
  object-centric models.
\newblock \emph{CoRR}, abs/2210.05861, 2022.

\bibitem[Xie et~al.(2022)Xie, Xie, and Zisserman]{xie2022segmenting}
Xie, J., Xie, W., and Zisserman, A.
\newblock Segmenting moving objects via an object-centric layered
  representation.
\newblock In \emph{NeurIPS}, 2022.

\bibitem[Yi \& Marshall(2000)Yi and Marshall]{yi00principal}
Yi, W. and Marshall, S.
\newblock Principal component analysis in application to object orientation.
\newblock \emph{Geo-spatial Information Science}, 2000.

\bibitem[Yu et~al.(2022)Yu, Guibas, and Wu]{yu22unsupervised}
Yu, H., Guibas, L.~J., and Wu, J.
\newblock Unsupervised discovery of object radiance fields.
\newblock In \emph{ICLR}, 2022.

\bibitem[Zhang et~al.(2022)Zhang, Li, Liu, Zhang, Su, Zhu, Ni, and
  Shum]{zhang22dino}
Zhang, H., Li, F., Liu, S., Zhang, L., Su, H., Zhu, J., Ni, L.~M., and Shum,
  H.-Y.
\newblock {{DINO}}: {{DETR}} with {{Improved DeNoising Anchor Boxes}} for
  {{End-to-End Object Detection}}, 2022.

\bibitem[Zhao et~al.(2020)Zhao, Jia, and Koltun]{zhao20exploring}
Zhao, H., Jia, J., and Koltun, V.
\newblock Exploring {{Self-Attention}} for {{Image Recognition}}.
\newblock In \emph{CVPR}, 2020.

\bibitem[Zhou et~al.(2022)Zhou, Zhang, Lee, Sun, Li, Zhu, Yoo, Qi, and
  Han]{zhou2021slot}
Zhou, Y., Zhang, H., Lee, H., Sun, S., Li, P., Zhu, Y., Yoo, B., Qi, X., and
  Han, J.-J.
\newblock {Slot-VPS}: Object-centric representation learning for video panoptic
  segmentation.
\newblock In \emph{CVPR}, 2022.

\bibitem[Zhu et~al.(2021)Zhu, Su, Lu, Li, Wang, and Dai]{zhu21deformable}
Zhu, X., Su, W., Lu, L., Li, B., Wang, X., and Dai, J.
\newblock Deformable {{DETR}}: {{Deformable Transformers}} for {{End-to-End
  Object Detection}}.
\newblock In \emph{ICLR}, 2021.

\end{thebibliography}
\bibliographystyle{icml2023}

\newpage
\appendix
\onecolumn

\section{Limitations}
\label{appendix:limitations}
Although invariant slot attention has many benefits, we have identified the following areas of improvement and possible future work.

\isatsr incorporates spatial symmetries in 2D pixel space whereas the world is 3D. Without prior knowledge of the object's underlying 3D geometry it is difficult to model, for example, out of plane rotation, shear, and non-rigid deformations.
Incorporating spatial symmetries in 3D extension of our method could help address this limitation.

Complex scenes are composed of not just objects but also ``stuff'' such as sky, road, and ocean. Modeling position, scale, and rotation of the sky, for example, is counter intuitive. Instead these could be modelled as background. A highly promising future direction is to combine Invariant Slot Attention with a background model -- do the foreground and background models require different symmetries?

A limitation, not of our method, but of our experiments is that we focus solely on object discovery whereas our architectural contribution is more general. Our position and scale invariant cross-attention approach could be, for example, ported to DETR for supervised object detection and panoptic segmentation by replacing Slot Attention slot queries with DETR object queries. As in recent versions of DETR, such as \citet{zhang22dino}, we use learnable initial slot positions and scales in some of our experiments. \citet{zhang22dino} predict the positions and scales of object queries (analogous to slots in slot attention), which are then refined throughout their decoder. We instead extract positions, rotations and scales from the attention masks to ensure equivariance of the computation.

Lastly, our experiments are still bottlenecked by the current capabilities of object discovery methods. As new methods for unsupervised / weakly-supervised instance level scene decomposition in in-the-wild data emerge, future work should investigate whether the explicit modeling of spatial symmetries continues to play a key role.

\section{Pseudocode}
\label{appendix:pseudocode}
The pseudo-code of \isatsr is provided in \cref{alg:isa}. This follows the discussion in \cref{sec:isa}. The only detail that we'd like to add here is a scaling parameter $\delta$ which scales \texttt{rel\_grid} in line 6. We set $\delta = 5$ in all our experiments. We do this for numerical and stability reasons. For example, this results in \texttt{rel\_grid} ranging from -2 to 2 instead of -10 to 10 for an object with $\slotscales = 0.1$.

\begin{algorithm*}
\caption{\textcolor{blue}{Translation}, \textcolor{gray}{Rotation}, and \textcolor{purple}{Scaling} Invariant Slot Attention.}
\label{alg:isa}

{\bfseries Inputs:} $\text{inputs} \in \mathbb{R}^{N{\times}D_{inputs}}$, \textcolor{blue}{$\text{abs\_grid} \in \mathbb{R}^{N{\times}2}$}, $\text{slots} \in \mathbb{R}^{K{\times}D_{slots}}$, \textcolor{blue}{Slot positions, $\slotpositions \in \mathbb{R}^{K{\times}2}$}, \\ \textcolor{gray}{Slot rotations, $\slotrotm \in \mathbb{R}^{K{\times}2{\times}2}$}, \textcolor{purple}{Slot scales, $\slotscales \in \mathbb{R}^{K{\times}2}$}, $T$ iterations, small $\epsilon$.

{\bfseries Data:} {Encoders $f, g, k, v, q$, parameters of LayerNorms, MLP and GRU, $\delta$.}

{\bfseries Outputs:}  {$\text{slots} \in \mathbb{R}^{K{\times}D_{slot}}$, \textcolor{blue}{$\slotpositions \in \mathbb{R}^{K{\times}2}$}, \textcolor{gray}{$\slotrotm \in \mathbb{R}^{K{\times}2{\times}2}$}, \textcolor{purple}{$\slotscales \in \mathbb{R}^{K{\times}2}$}.}

\begin{algorithmic}[1]

\STATE{$\text{inputs} = \text{LayerNorm}(\text{inputs})$}
\FOR{$t = 1$ {\bfseries to} $T$ \textcolor{blue}{$+\ 1$}}
    \STATE{$\text{slots\_prev} = \text{slots}$}
    \STATE{$\text{slots} = \text{LayerNorm}(\text{slots})$}
    \item[]
    \STATE{\textcolor{darkgreen}{\# Computes relative grids per slot, and associated key, value embeddings.}}
    \FOR{$k = 1$ {\bfseries to} $K$}
        \STATE{\textcolor{blue}{$\text{rel\_grid}^k = [\textcolor{gray}{(\slotrotm^k)^{-1}}(\text{abs\_grid} - \slotpositions^k)]$} \textcolor{purple}{$/\ \left(\slotscales^k \times \delta\right)$}}
        \STATE{\textcolor{blue}{$\text{keys}^k = f\left(k(\text{inputs}) + g(\text{rel\_grid}^k)\right)$}}
        \STATE{\textcolor{blue}{$\text{values}^k = f\left(v(\text{inputs}) + g(\text{rel\_grid}^k)\right)$}}
    \ENDFOR
    \item[]
    \STATE{\textcolor{darkgreen}{\# Inverted dot production attention.}}
    \FOR{$k = 1$ {\bfseries to} $K$}
        \STATE{$\text{attn}^k = \frac{1}{\sqrt{K}} \text{keys}^k * q(\text{slots}^k)^T$}
    \ENDFOR
    \STATE{$\text{attn} = \text{softmax}(\text{attn}, \text{axis}=\text{``slots''})$}
    \STATE{$\text{updates} = \text{WeightedMean}(\text{weights}=\text{attn}, \text{values}=\text{values})$}
    \STATE{\textcolor{blue}{$\text{attn} \mathrel{/}= \text{Sum}(\text{attn}, \text{axis}=\text{``inputs''})$}}
    \item[]
    \STATE{\textcolor{darkgreen}{\# Updates $\slotpositions$, $\slotscales$ and $\text{slots}$.}}
    \FOR{$k = 1$ {\bfseries to} $K$}
    \STATE{\textcolor{blue}{$\slotpositions^k = \text{WeightedMean}(\text{weights}=\text{attn}^k, \text{values}=\text{abs\_grid})$}}
    \STATE{\textcolor{gray}{$\slotrotm^k = \text{Symmetrize}(\text{WeightedPCAAnalytical}(\text{inputs}=\text{abs\_grid} - \slotpositions^k, \text{weights}=\text{attn}^k)$}}
    \STATE{\textcolor{purple}{$\slotscales^k = \sqrt{\text{WeightedMean}(\text{weights}=\text{attn}^k + \epsilon, \text{values}=[\textcolor{gray}{(\slotrotm^k)^{-1}}(\text{abs\_grid} - \slotpositions^k)]^2)}$}}
    \ENDFOR
    \IF{\textcolor{blue}{$t < T + 1$}}
        \STATE{$\text{slots} = \text{GRU}(\text{state}=\text{slots\_prev}, \text{inputs}=\text{updates})$}
        \STATE{\text{slots} $\mathrel{{+}{=}}$ \text{MLP}(\text{LayerNorm}(\text{slots}))}
    \ENDIF
\ENDFOR

\end{algorithmic}
\end{algorithm*}

\section{Model details}
\label{appendix:model}
\subsection{Architecture details}
\label{appendix:model:arch}
The model consists of the same three components as Slot Attention: an encoder $E_{\theta}$, the attention module, and the decoder $D_{\phi}$. In our experiments, two types of visual backbones were used: 1) a shallow CNN encoder similar to that of \cite{locatello20object} and 2) a  ResNet-34~\citep{he2016deep}, details of which are explained below. In all cases the spatial broadcast operation in the SB decoder matched the number of tokens output by the encoder.

\paragraph{Encoder details}
For the Tetrominoes dataset, the shallow CNN encoder consists of 4 convolutional layers (64 channels, kernel size $5\times 5$, stride 1, `SAME' padding) and ReLU activations \cite{nair2010rectified}. The output feature map size is $35\times 35$ which is the same as the input size.

In the Objects Room dataset, the shallow CNN encoder is the same as that in Tetrominoes except that the first two layers have stride 2 and that input images are of size $64\times64$ resulting in feature maps of size $16\times16$.

For all other synthetic datasets (CLEVR, CLEVRTex, MultishapeNet), the shallow CNN encoder consists of 4 convolutional layers (64 channels, kernel size $5\times5$, `SAME' padding) and ReLU activations. The stride is 2 for the first three layers and 1 for the remaining one layer. Input images are of size $128\times 128$. We thus obtain feature maps of size $16\times16$.

ResNet-34 \cite{he2016deep} is the standard ResNet model with 34 convolutional layers. We used group normalization \cite{wu2018group} instead of batch normalization \cite{ioffe2015batch} to avoid a dependence on training batch size. All experiments that used this encoder, as opposed to the shallow CNN, have been marked with `(ResNet)' in the main manuscript. This includes several experiments on the CLEVRTex dataset and all the experiments on the WaymoOpen dataset. The inspiration to use a bigger backbone on more complex data comes from the SAVi++ method \cite{elsayed22savipp}. Input images are $128\times 128$ for CLEVRTex and $128\times192$ for Waymo Open. We reduce the stride in the ResNet root block ($3\times3$ convolutional layer in the root block with stride 1 with no max pooling operation after it, as is conventional for ResNets on small input images). Thus the output feature map size is $16\times16$ for CLEVRTex and $16\times24$ for Waymo Open.

\paragraph{Attention module details}
The attention module consists of functions $\mathcal{Q}, \mathcal{K}, \mathcal{V},~\mathrm{and}~g$ which are linear projections (without bias, except for $g$ which has a bias term) and function $f$ which is a MLP. The linear projections have output dimension $64$, the MLP has a hidden size $128$, output dimension $64$, and applies layer-norm on its inputs (``pre-norm''). We used three iterations of slot attention as in the original publication \cite{locatello20object} which are mediated by a Gated Recurrent Unit and a MLP with hidden size $128$, pre-normalization, output size $64$, and a residual connection.

All experiments use 11 slots except in Tetrominoes where we use 4 slots and MultiShapeNet where we use 5 slots in both splits. Slots in slot attention are iteratively updated but must be initialized. We used a learned initialization in all cases. The learnable weights are themselves initialized using a standard normal distribution $\mathcal{N}(0, 1)$. Reference frames associated with these slots are also initialized with learnable embeddings in all experiments except in Tetrominoes and in CLEVR where they are randomly sampled in every forward pass. In those experiments where they are learnable embeddings, $\slotpositions$ are initialized using $\mathcal{U}\left[-1, 1\right]$, $\slotscales$ are initialized using $\mathcal{N}(0.1, 0.01)$, $S_r$ are initialized using $\frac{\pi}{4} \mathrm{tanh}\left( \mathcal{N}(0., 0.1) \right)$. In those experiments where they are randomly sampled in every forward pass, $\slotpositions$ are sampled from $\mathcal{U}\left[-1, 1\right]$, $\slotscales$ are sampled from $\mathcal{N}(0.1, 0.1)$, and $S_r$ are sampled from $\mathcal{U}\left[-\frac{\pi}{4}, \frac{\pi}{4}\right]$.

\paragraph{Spatial broadcast decoder details}
In Tetrominoes, we use an MLP that decodes each pixel independently given the slot vector and pixel coordinates. It has five layers, with hidden size 256, interleaved with ReLU activations.

For the Objects Room dataset, we use a CNN decoder with 5 transpose convolutional layers (kernel size $5 \times 5$, stride 2 for the first two layers and stride 1 for the remaining, `SAME' padding, 64 channels) interleaved by ReLU activations. Input feature maps are spatially broadcasted position encoded slots at a resolution of $16\times16$. The output image is of size $64 \times 64$.

All other datasets use the same setup as the Objects Room dataset, except that outputs are of size $128\times 128$ (except Waymo Open at $128\times 192$) and so while input feature maps are at $16\times16$ ($16\times24$ for Waymo Open) there is one extra convolution transpose layer with stride 2.

A final dense layer (also referred to as a $1\times 1$ conv.) projects the decoder output to the 3 channel RGB predictions.

Function $h$ is applied to \texttt{rel\_grid} and is a linear projection to $64$ dimensions with a bias term.

\subsection{Rotation estimation}
\label{appendix:model:rot}
Our heuristic assumes that the orientation of a slot is given by the axis with the highest variation (the first principal component). E.g., the axis of an outline of a car would be horizontal, whereas the axis of an outline of a person would be vertical. We can compute this axis $v_1^k$ (and a second orthogonal axis $v_2^k$) for each slot $k$ by using weighted Principal Component Analysis (PCA) with $\mathrm{abs\_grid}$ as the input and $\mathrm{attn}$ as the weights:
\begin{align}
    v_1^k, v_2^k &= \mathrm{WeightedPCA}(\mathrm{abs\_grid}, \mathrm{attn}^k)
\end{align}
Importantly, we can compute the axes \textit{analytically} by computing the eigenvalues of a $2{\times}2$ weighted covariance matrix. This facilitates stable gradients when we backpropagate through the rotation detection. We further `post-process'
the axes to (1) make sure the coordinate grid is always left-handed (so that we do not accidentally mirror objects) and (2) limit rotations to $[0, 45^{\circ}]$. The latter partially accounts for the ambiguity in the detected rotation -- our heuristic attempts to align the object either horizontally or vertically with a gentle rotation without flipping it upside down, etc.:
\begin{align}
    \tilde{v}_1^k, \tilde{v}_2^k &= \textrm{post-process}\left(v_1^k, v_2^k\right)\,, \\
    S^k_r &= \begin{bmatrix} \vertbar & \vertbar \\ \tilde{v}_1^k & \tilde{v}_2^k \\ \vertbar & \vertbar \\\end{bmatrix}\,.
\end{align}
$\tilde{v}_1$ and $\tilde{v}_2$ form the columns of a rotation matrix $S^k_r$.

\section{Optimization details}
The model is trained using Adam \citep{kingma15adam} with a learning rate of $4\times10^{-4}$ on all datasets except for Waymo Open Depths, where we use $2\times10^{-4}$ following \cite{elsayed22savipp}. We use a learning rate warm-up going from 0 for 50k steps. Afterwards, the learning rate decays using cosine decay to 0 \citep{loshchilov17sgdr}. We train for 500k training steps, except for Waymo, 300k, and Tetrominoes, 50k. In Tetrominoes, we use 5k steps warm-up. The batch size is 64.

\section{Evaluation metrics}
\label{appendix:metrics}
In all datasets, except for Objects Room where we report ARI, we report the foreground adjusted rand index (FG-ARI). This metric measures the foreground object decomposition and is based on the ARI metric \cite{hubert1985comparing} popular in clustering literature. It measures, for each pair of pixels, whether their grouping is the same in the predicted segmentation maps and in the ground truth. It is permutation invariant and therefore suitable for class agnostic instance segmentation without the need for explicit matching between predicted and ground truth segments. FG-ARI has been reported in several recent works in the community such as \cite{singh22simple,kipf22conditional,locatello20object}.

We also report image reconstruction performance in terms of mean squared error (MSE) in some datasets. These are reported mainly for completeness and the model has not been optimized to produce high quality reconstructions.

\section{Dataset details}
\label{appendix:dataset_details}

\subsection{Waymo Open}
For depth prediction experiments, we use a pre-trained Dense Prediction Transformer (DPT, \citet{ranftl21vision}) to predict disparity masks for all images, which we then normalize to a 0-1 range \textit{per image}. The DPT prediction are then used as \textit{targets} for the Slot Attention decoder. 

\section{Additional experimental results}
\label{appendix:all_results}

\subsection{Invariance to rotations}
All quantitative results comparing \isats and \isatsr are reported in \cref{app:tab:rot}.
\begin{table}[h]
\centering
\caption{\textbf{Rotation invariance}: Comparing \isats\; against \isatsr\; in various benchmarks. Objects room results are ARIs whereas all others are FG-ARIs.}
\vspace{2mm}
\begin{tabular}{lllll}
    \toprule
                          & \multicolumn{2}{c}{(FG-)ARI $\uparrow$}  \\
    \textbf{Dataset}      & \isats  & \isatsr \\
    \midrule
    Objects room (w/ bg)  &          &          \\
    Validation            & \ebar{85.5}{6.6} & \ebar{84.3}{4.6}\\
    Six Objects           & \ebar{84.5}{4.6} & \ebar{83.2}{2.7} \\
    Empty Room            & \ebar{83.6}{8.8} & \ebar{81.5}{7.1} \\
    Identical Colors      & \ebar{85.1}{5.9} & \ebar{83.8}{3.9} \\
    \midrule
    CLEVR                 & \ebar{98.9}{0.2} & \ebar{98.0}{0.9} \\
    \midrule
    MultiShapeNet    &          &          \\
    - All Data            & \ebar{69.8}{1.1} & \ebar{77.7}{5.5} \\
    - Four Objects        & \ebar{86.5}{1.1} & \ebar{80.7}{6.4} \\
    \midrule
    CLEVRTex (CNN)              & \ebar{78.8}{3.9} & \ebar{79.6}{5.5} \\
    CLEVRTex CAMO (CNN)        & \ebar{72.9}{3.5} & \ebar{73.8}{4.9} \\
    CLEVRTex OOD (CNN)          & \ebar{73.2}{3.1} & \ebar{74.9}{3.8} \\
    \midrule
    CLEVRTex (ResNet)              & \ebar{92.9}{0.4} & \ebar{93.3}{0.7} \\
    CLEVRTex CAMO (ResNet)        & \ebar{86.2}{0.8} & \ebar{87.0}{1.7} \\
    CLEVRTex OOD (ResNet)          & \ebar{84.4}{0.8} & \ebar{84.9}{1.2} \\
   \bottomrule
\end{tabular}
\label{app:tab:rot}
\end{table}

\subsection{CLEVR dataset}
A table of results for the CLEVR dataset are presented in Table \ref{tab:clevr}.
\begin{table}[t!]
\centering
\caption{Results on the CLEVR dataset. Except for AST which used 3 seeds, all methods used 5 seeds. AST-Seg-B3-CT numbers are copied from \citet{sauvalle22unsupervised}.}
\vspace{2mm}
    \begin{tabular}{lll}
        \toprule
        \textbf{Method} & $\uparrow$\textbf{FG-ARI} & $\downarrow$\textbf{MSE} \\
        \midrule
        \begin{filecontents*}{csv/clevr.tex}
method,fg_ari_mean,fg_ari_std,mse_mean,mse_std
AST-Seg-B3-CT,98.3,0.1,16,1
SA,98.8,0.2,13,2
\isat,\textbf{99.0},0.2,\textbf{11},2
\isats,\underline{98.9},0.2,\textbf{11},1
\isatsr,98.0,0.9,12,2
\end{filecontents*}
\csvreader[
                head to column names,
                late after line=\\
            ]{csv/clevr.tex}{1=\one, 2=\two, 3=\three, 4=\four, 5=\five}
            {\one & {\two}{\color{lightgrey}\tiny$\pm$\three} & {\four}{\color{lightgrey}\tiny$\pm$\five}}
        \bottomrule
    \end{tabular}
    \label{tab:clevr}
\end{table}

\subsection{Objects Room dataset}
A table version of the bar plot in \cref{fig:msneplusobjsrm}-Top, is presented in \cref{tab:objects_room}.
\begin{table*}[t]
\centering
\caption{Result on Objects Room with background segments included, 10 random seeds.}
\vspace{2mm}
\begin{tabular}{lllllllll}
    \toprule
    \textbf{Method} & \multicolumn{4}{c}{$\uparrow$\textbf{ARI (11 slots), background included.}} \\
     & \textbf{Validation} & \textbf{Six Objects} & \textbf{Empty Room} & \textbf{Identical Colors} \\
    \midrule
    \begin{filecontents*}{csv/objects_room.tex}
method,val,val_sd,test_six,test_six_sd,test_empty,test_empty_sd,test_identical,test_identical_sd
SA,63.0,17.9,62.8,18.4,58.4,16.4,61.2,17.8
\isat,83.5,9.2,82.5,6.7,78.7,14.3,82.4,8.2
\isats,\textbf{85.5},6.6,\textbf{84.5},4.6,\textbf{83.6},8.8,\textbf{85.1},5.9
\isatsr,\underline{84.3},4.6,\underline{83.2},2.7,\underline{81.5},7.1,\underline{83.8},3.9
\end{filecontents*}
\csvreader[
        head to column names,
        late after line=\\
    ]{csv/objects_room.tex}{1=\one, 2=\two, 3=\three, 4=\four, 5=\five, 6=\six, 7=\seven, 8=\eight, 9=\nine}
    {\one & {\two}{\color{lightgrey}\tiny$\pm$\three} & {\four}{\color{lightgrey}\tiny$\pm$\five} & {\six}{\color{lightgrey}\tiny$\pm$\seven} & {\eight}{\color{lightgrey}\tiny$\pm$\nine}}
    \bottomrule
\end{tabular}
\label{tab:objects_room}
\end{table*}

\subsection{MultiShapeNet}
A table version of the bar plot in \cref{fig:msneplusobjsrm}-Bottom, along with the performance of \isatsr, is presented in \cref{tab:msne}. We further show learned segmentation masks, slot positions and scales in Figure \ref{fig:msne_frames_fig}.

\begin{table}[h]
\centering
\caption{Results on MultiShapeNet-Easy, 5 random seeds.}
\vspace{2mm}
    \begin{tabular}{lllll}
            \toprule
            \textbf{Method} & \multicolumn{2}{c}{\textbf{All Data}} & \multicolumn{2}{c}{\textbf{Four Objects}} \\
             & $\uparrow$\textbf{FG-ARI} & $\downarrow$\textbf{MSE} & $\uparrow$\textbf{FG-ARI} & $\downarrow$\textbf{MSE} \\
            \midrule
            \begin{filecontents*}{csv/msne.tex}
method,all_fg_ari_mean,all_fg_ari_std,all_mse_mean,all_mse_std,four_fg_ari_mean,four_fg_ari_std,four_mse_mean,four_mse_std
SA,56.8,2.6,33,5,71.5,11.6,50,2
\isat,\underline{71.2},3.5,\underline{29},3,\underline{82.3},5.3,\underline{43},4
\isats,69.8,1.1,\textbf{27},1,\textbf{86.5},1.1,\textbf{41},3
\isatsr,\textbf{77.7},5.5,35,3,80.7,6.4,51,6
\end{filecontents*}
\csvreader[
                column count=6,
                head to column names,
                late after line=\\
            ]{csv/msne.tex}{1=\one, 2=\two, 3=\three, 4=\four, 5=\five, 6=\six, 7=\seven, 8=\eight, 9=\nine}
            {\one & {\two}{\color{lightgrey}\tiny$\pm$\three} & {\four}{\color{lightgrey}\tiny$\pm$\five} & {\six}{\color{lightgrey}\tiny$\pm$\seven} & {\eight}{\color{lightgrey}\tiny$\pm$\nine}}
    \bottomrule
    \end{tabular}
    \label{tab:msne}
\end{table}

\newcommand{\sizeframesmsne}{110pt}
\begin{figure*}[t]
    \centering
    \includegraphics[height=\sizeframesmsne]{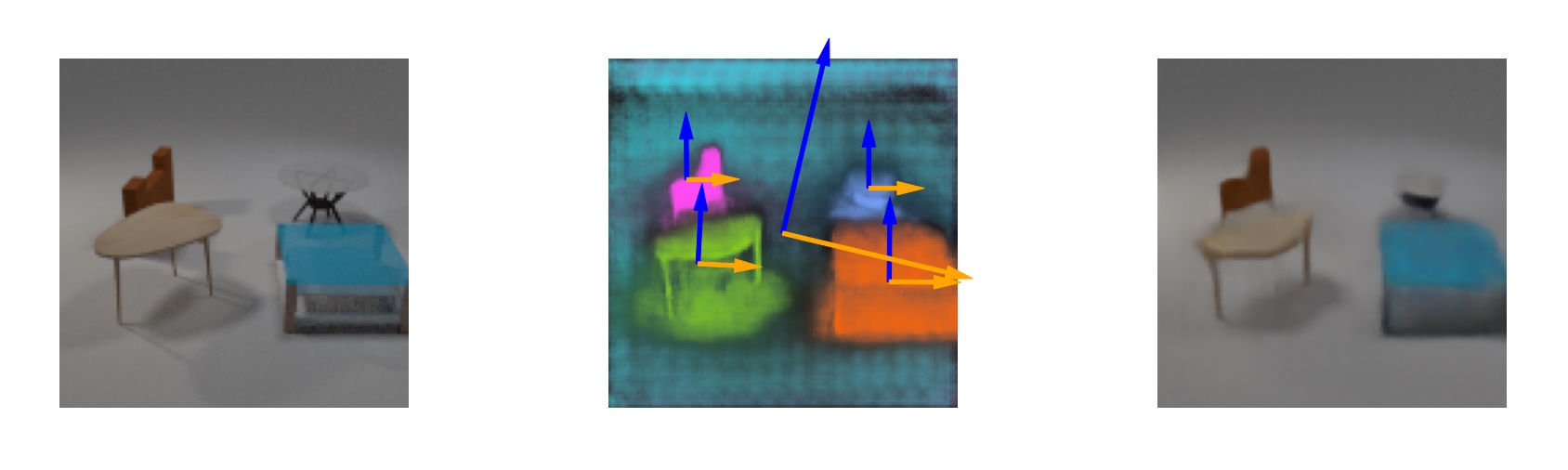}
    \includegraphics[height=\sizeframesmsne]{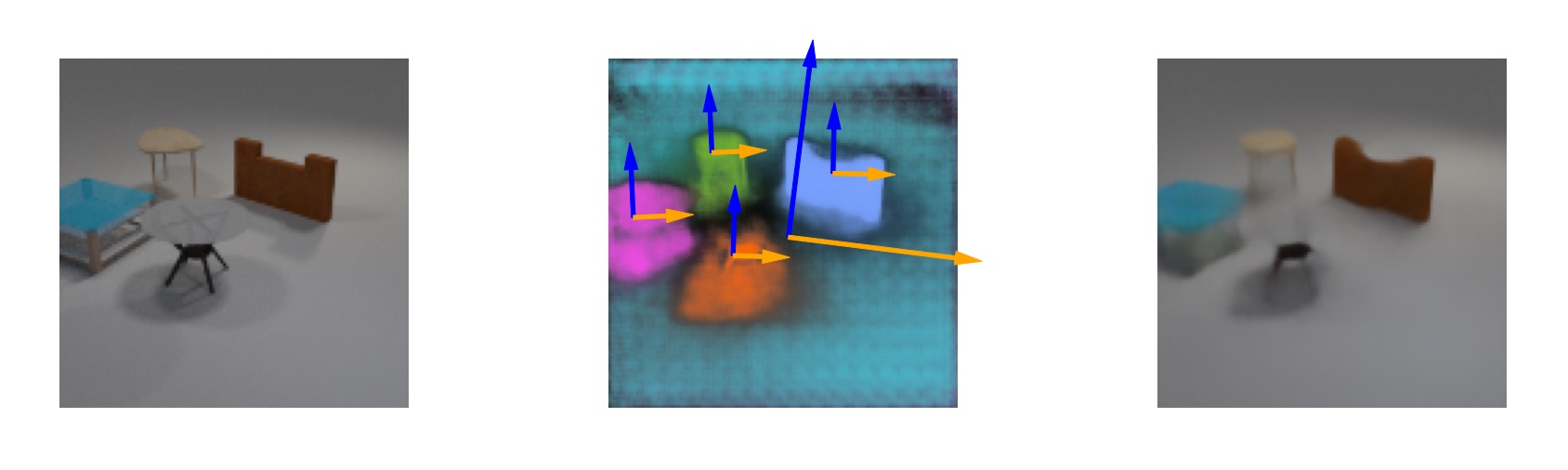}
    \includegraphics[height=\sizeframesmsne]{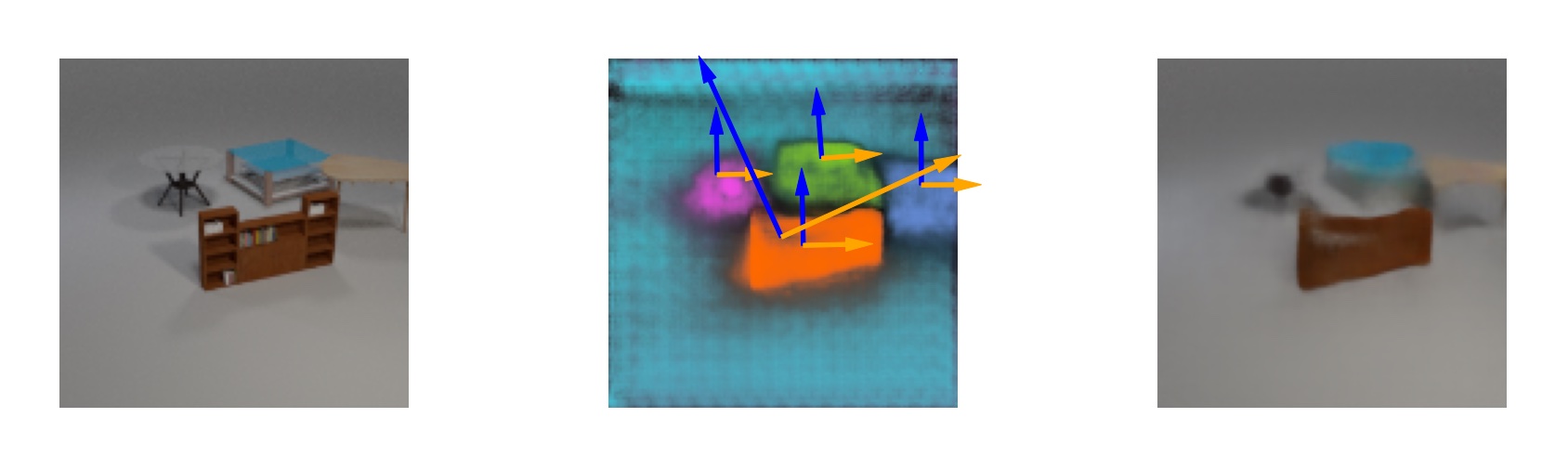}
    \includegraphics[height=\sizeframesmsne]{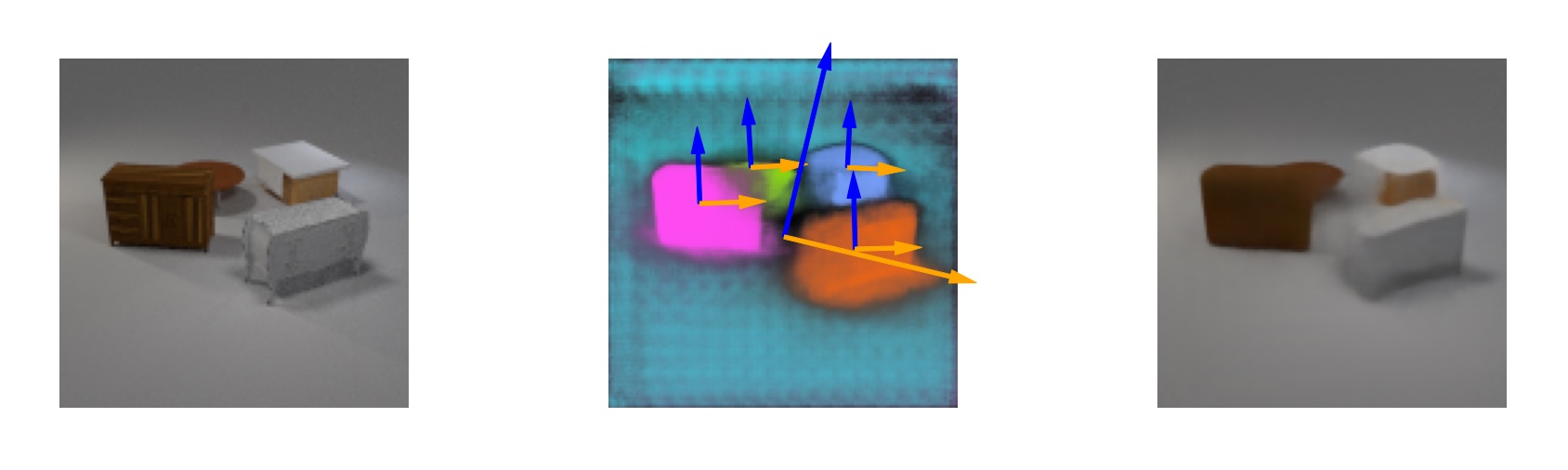}
    \includegraphics[height=\sizeframesmsne]{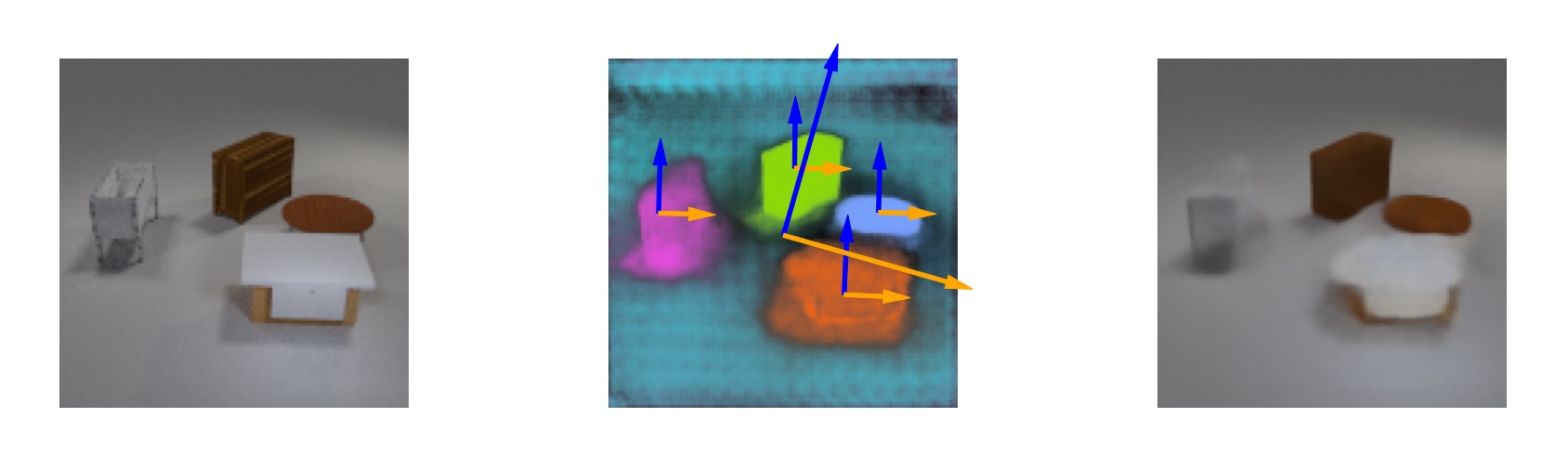}
    \caption{Per-slot reference frames learned without supervision on MultiShapeNet. Left column: input image. Right column: reconstructed image. Middle column: we show the (x, y) position, (sx, sy) scale and orientation of individual slots overlaid over a predicted soft segmentation mask. Each pair of blue and orange arrows corresponds to one slot. The longer the arrows, the higher the scale of the slot is. We filter out slots that have a low predicted probability in the segmentation mask (i.e. they are probably inactive).}
    \label{fig:msne_frames_fig}
\end{figure*}

\subsection{CLEVRTex results}
\begin{figure*}[t]
    \centering
    \includegraphics[width=\textwidth]{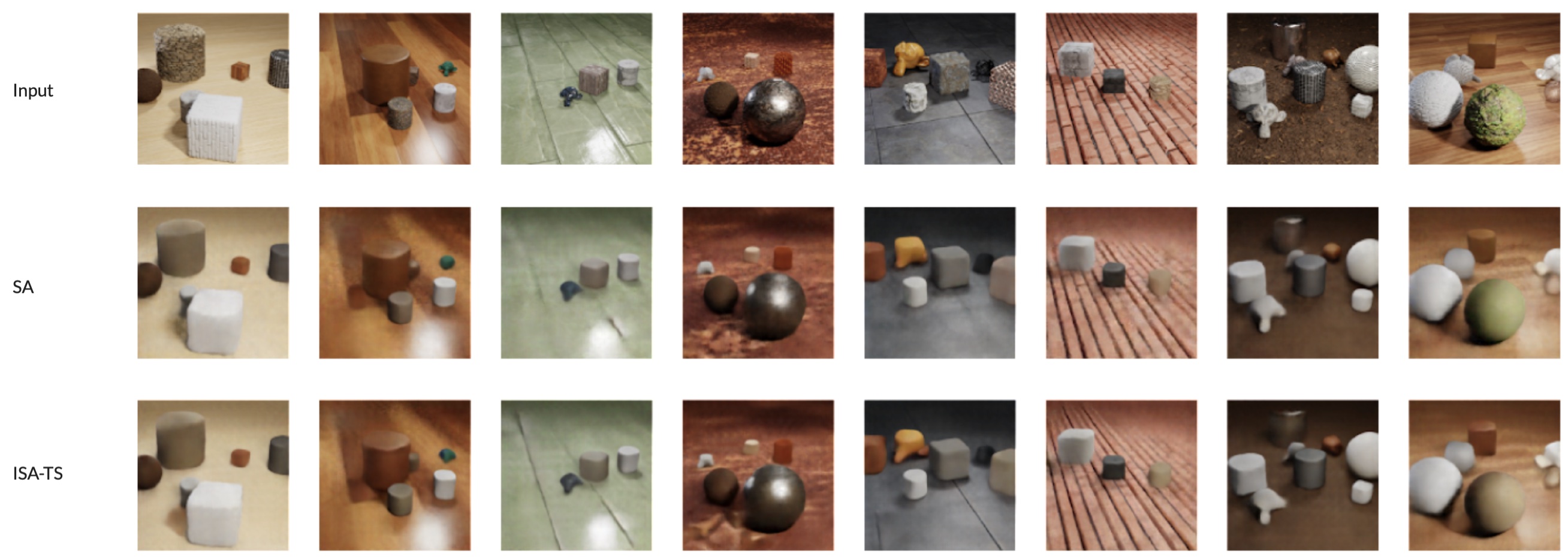}
    \caption{Comparison between Slot Attention and Translation and Scale Invariant Slot Attention in image decoding.}
    \label{fig:clevrtex_rec}
\end{figure*}

Table \ref{tab:clevrtex:full} compiles all evaluations on the CLEVRTex alongside FG-ARI and MSE metrics and several baselines. We further report FG-mIoU results in Table \ref{tab:clevrtex:miou}. We further show image decodings, learned segmentation masks, slot positions, orientations and scales in Figures \ref{fig:clevrtex_frames_fig}, \ref{fig:clevrtex_segm} and \ref{fig:clevrtex_rec}.

\begin{table*}[h]
    \centering
    \caption{CLEVRTex results on the test set, CAMO set (objects and backgrounds blend together) and OOD set (novel textures). Prior results taken from \citep{karazija21clevrtex} use 3 random seeds, we use 10 random seeds. FG-ARI is reported in \%.}
    \vspace{2mm}
    \begin{tabular}{lllllll}
        \toprule
        \textbf{Method} & \multicolumn{2}{c}{\textbf{CLEVRTex}} & \multicolumn{2}{c}{\textbf{CLEVRTex CAMO}} & \multicolumn{2}{c}{\textbf{CLEVRTex OOD}} \\
         & $\uparrow$FG-ARI & $\downarrow$MSE & $\uparrow$FG-ARI & $\downarrow$MSE & $\uparrow$FG-ARI & $\downarrow$MSE \\
        \midrule
        \begin{filecontents*}{csv/clevrtex_baselines_2.tex}
method,test_miou,test_miou_sd,test_fg_ari,test_fg_ari_sd,test_mse,test_mse_sd,camo_miou,camo_miou_sd,camo_fg_ari,camo_fg_ari_sd,camo_mse,camo_mse_sd,ood_miou,ood_miou_sd,ood_fg_ari,ood_fg_ari_sd,ood_mse,ood_mse_sd
SPACE,9.1,3.5,17.5,4.1,298,80,8.7,3.5,10.6,2.1,251,61,6.9,3.3,12.7,3.4,\textbf{387},66
DTI,33.8,1.3,79.9,1.4,438,22,27.5,1.6,72.9,1.9,377,17,32.6,1.1,73.7,1.0,590,4
Gen-V2,7.9,1.5,31.2,12.4,315,106,7.5,1.7,29.6,12.8,278,75,8.7,1.6,29.0,11.2,539,147
eMORL,12.6,2.4,45.0,7.8,318,43,11.6,2.1,42.3,7.2,269,31,13.2,2.6,43.1,9.3,471,51
AST-Seg-B3-CT,79.6,0.5,\textbf{94.8},0.5,\textbf{139},7,73.1,0.7,\underline{87.3},3.8,\textbf{145},6,67.5,0.8,83.1,0.8,832,24
\end{filecontents*}
\csvreader[
            column count=11,
            head to column names,
            late after line=\\
        ]{csv/clevrtex_baselines_2.tex}{1=\one, 4=\two, 5=\three, 6=\four, 7=\five, 10=\six, 11=\seven, 12=\eight, 13=\nine, 16=\ten, 17=\eleven, 18=\twelve, 19=\thirteen}
        {\one & {\two}{\color{lightgrey}\tiny$\pm$\three} & {\four}{\color{lightgrey}\tiny$\pm$\five} & {\six}{\color{lightgrey}\tiny$\pm$\seven} & {\eight}{\color{lightgrey}\tiny$\pm$\nine} & {\ten}{\color{lightgrey}\tiny$\pm$\eleven} & {\twelve}{\color{lightgrey}\tiny$\pm$\thirteen}}
        \midrule
        \begin{filecontents*}{csv/clevrtex_simplecnn_2.tex}
method,test_miou,test_miou_sd,test_fg_ari,test_fg_ari_sd,test_mse,test_mse_sd,camo_miou,camo_miou_sd,camo_fg_ari,camo_fg_ari_sd,camo_mse,camo_mse_sd,ood_miou,ood_miou_sd,ood_fg_ari,ood_fg_ari_sd,ood_mse,ood_mse_sd
SA,-,-,54.5,1.6,241.2,14.0,-,-,53.0,1.6,216.6,11.8,-,-,54.2,2.6,\underline{414.7},27.7
\isat (CNN),-,-,66.8,5.7,229.8,20.4,-,-,65.0,4.9,213.2,16.3,-,-,65.1,4.8,458.6,25.2
\isats (CNN),-,-,78.8,3.9,223.8,6.5,-,-,72.9,3.5,210.9,6.2,-,-,73.2,3.1,480.6,32.6
\isatsr (CNN),-,-,79.6,5.5,234.5,6.6,-,-,73.8,4.9,220.5,8.8,-,-,74.9,3.8,499.9,56.9
+ append,-,-,85.4,2.4,234.2,3.9,-,-,78.8,2.2,223.5,4.6,-,-,78.2,1.9,554.0,25.1
\end{filecontents*}
\csvreader[
            column count=11,
            head to column names,
            late after line=\\
        ]{csv/clevrtex_simplecnn_2.tex}{1=\one, 4=\two, 5=\three, 6=\four, 7=\five, 10=\six, 11=\seven, 12=\eight, 13=\nine, 16=\ten, 17=\eleven, 18=\twelve, 19=\thirteen}
        {\one & {\two}{\color{lightgrey}\tiny$\pm$\three} & {\four}{\color{lightgrey}\tiny$\pm$\five} & {\six}{\color{lightgrey}\tiny$\pm$\seven} & {\eight}{\color{lightgrey}\tiny$\pm$\nine} & {\ten}{\color{lightgrey}\tiny$\pm$\eleven} & {\twelve}{\color{lightgrey}\tiny$\pm$\thirteen}}
        \midrule
        \begin{filecontents*}{csv/clevrtex_resnet_2.tex}
method,test_miou,test_miou_sd,test_fg_ari,test_fg_ari_sd,test_mse,test_mse_sd,camo_miou,camo_miou_sd,camo_fg_ari,camo_fg_ari_sd,camo_mse,camo_mse_sd,ood_miou,ood_miou_sd,ood_fg_ari,ood_fg_ari_sd,ood_mse,ood_mse_sd
SA (ResNet),-,-,91.3,2.7,206.7,20.8,-,-,84.9,2.9,224.4,21.9,-,-,81.4,1.4,629.1,29.9
\isat (ResNet),-,-,87.4,6.6,197.1,27.0,-,-,79.0,5.9,217.6,24.5,-,-,78.6,4.9,548.1,21.3
\isats (ResNet),-,-,92.9,0.4,176.9,3.4,-,-,86.2,0.8,196.3,7.1,-,-,84.4,0.8,578.3,22.5
\isatsr (ResNet),-,-,93.3,0.7,185.9,6.4,-,-,87.0,1.7,206.1,10.0,-,-,\textbf{84.9},1.2,595.2,35.6
+ append,-,-,\underline{93.6},0.8,\underline{185.6},5.8,-,-,\textbf{87.5},0.9,\underline{201.1},6.9,-,-,\underline{84.8},0.7,647.1,24.6
- dec. only,-,-,93.4,1.0,200.4,6.6,-,-,87.2,1.7,223.5,5.3,-,-,83.9,1.6,614.4,49.7
- stop grad,-,-,73.4,10.5,392.6,116.5,-,-,62.8,10.3,388.2,81.2,-,-,67.5,8.0,621.9,62.8
\end{filecontents*}
\csvreader[
            column count=11,
            head to column names,
            late after line=\\
        ]{csv/clevrtex_resnet_2.tex}{1=\one, 4=\two, 5=\three, 6=\four, 7=\five, 10=\six, 11=\seven, 12=\eight, 13=\nine, 16=\ten, 17=\eleven, 18=\twelve, 19=\thirteen}
        {\one & {\two}{\color{lightgrey}\tiny$\pm$\three} & {\four}{\color{lightgrey}\tiny$\pm$\five} & {\six}{\color{lightgrey}\tiny$\pm$\seven} & {\eight}{\color{lightgrey}\tiny$\pm$\nine} & {\ten}{\color{lightgrey}\tiny$\pm$\eleven} & {\twelve}{\color{lightgrey}\tiny$\pm$\thirteen}}
        \bottomrule
    \end{tabular}
    \label{tab:clevrtex:full}
\end{table*}

\begin{table*}[h]
    \centering
    \caption{CLEVRTex results on the test set, CAMO set (objects and backgrounds blend together) and OOD set (novel textures). FG-mIoU is reported in \%. 10 seeds.}
    \vspace{2mm}
    \begin{tabular}{lllllll}
        \toprule
        \textbf{Method} & \multicolumn{2}{c}{\textbf{CLEVRTex}} & \multicolumn{2}{c}{\textbf{CLEVRTex CAMO}} & \multicolumn{2}{c}{\textbf{CLEVRTex OOD}} \\
         & $\uparrow$FG-mIoU & $\downarrow$MSE & $\uparrow$FG-mIoU & $\downarrow$MSE & $\uparrow$FG-mIoU & $\downarrow$MSE \\
        \midrule
        \begin{filecontents*}{csv/clevrtex_resnet_miou.tex}
method,test_miou,test_miou_sd,test_fg_ari,test_fg_ari_sd,test_mse,test_mse_sd,camo_miou,camo_miou_sd,camo_fg_ari,camo_fg_ari_sd,camo_mse,camo_mse_sd,ood_miou,ood_miou_sd,ood_fg_ari,ood_fg_ari_sd,ood_mse,ood_mse_sd
SA (ResNet),69.7,2.2,91.3,2.7,206.7,20.8,63.9,2.1,84.9,2.9,224.4,21.9,63.5,1.6,81.4,1.4,629.1,29.9
\isat (ResNet),66.4,3.2,87.4,6.6,197.1,27.0,58.7,2.7,79.0,5.9,217.6,24.5,61.7,2.8,78.6,4.9,\textbf{548.1},21.3
\isats (ResNet),\textbf{72.4},0.8,92.9,0.4,\textbf{176.9},3.4,\textbf{65.6},0.6,86.2,0.8,\textbf{196.3},7.1,\textbf{66.7},0.9,84.4,0.8,578.3,22.5
\isatsr (ResNet),71.4,1.5,93.3,0.7,185.9,6.4,64.9,1.2,87.0,1.7,206.1,10.0,66.3,1.3,\textbf{84.9},1.2,595.2,35.6
\end{filecontents*}
\csvreader[
            column count=11,
            head to column names,
            late after line=\\
        ]{csv/clevrtex_resnet_miou.tex}{1=\one, 2=\two, 3=\three, 6=\four, 7=\five, 8=\six, 9=\seven, 12=\eight, 13=\nine, 14=\ten, 15=\eleven, 18=\twelve, 19=\thirteen}
        {\one & {\two}{\color{lightgrey}\tiny$\pm$\three} & {\four}{\color{lightgrey}\tiny$\pm$\five} & {\six}{\color{lightgrey}\tiny$\pm$\seven} & {\eight}{\color{lightgrey}\tiny$\pm$\nine} & {\ten}{\color{lightgrey}\tiny$\pm$\eleven} & {\twelve}{\color{lightgrey}\tiny$\pm$\thirteen}}
        \bottomrule
    \end{tabular}
    \label{tab:clevrtex:miou}
\end{table*}

\newcommand{\sizeframes}{110pt}
\begin{figure*}[t]
    \centering
    \includegraphics[height=\sizeframes]{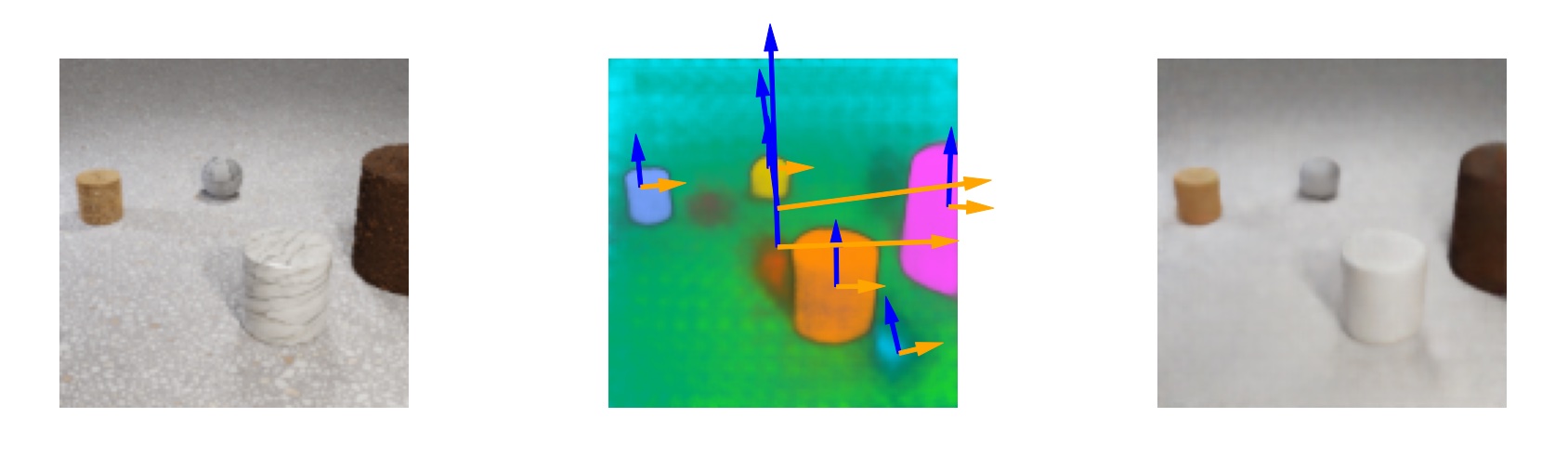}
    \includegraphics[height=\sizeframes]{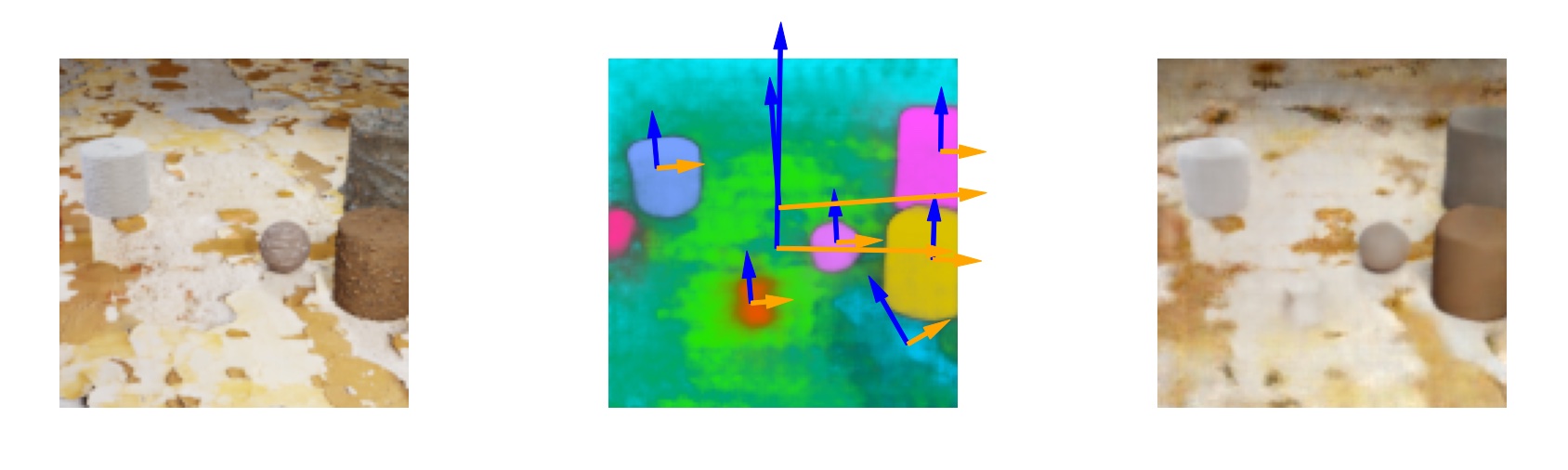}
    \includegraphics[height=\sizeframes]{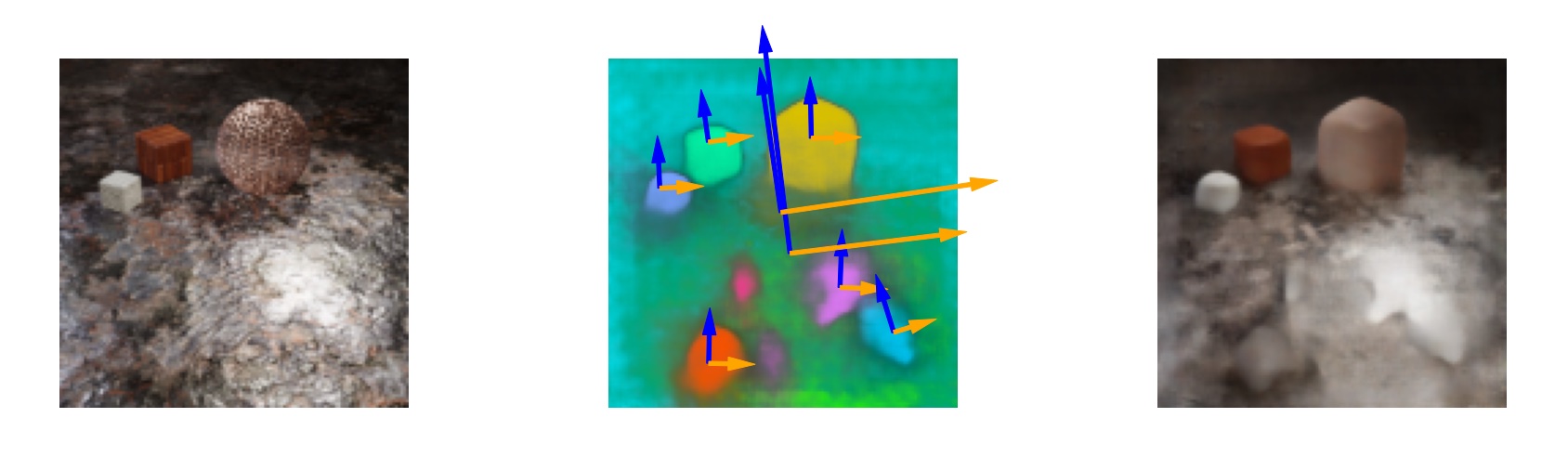}
    \includegraphics[height=\sizeframes]{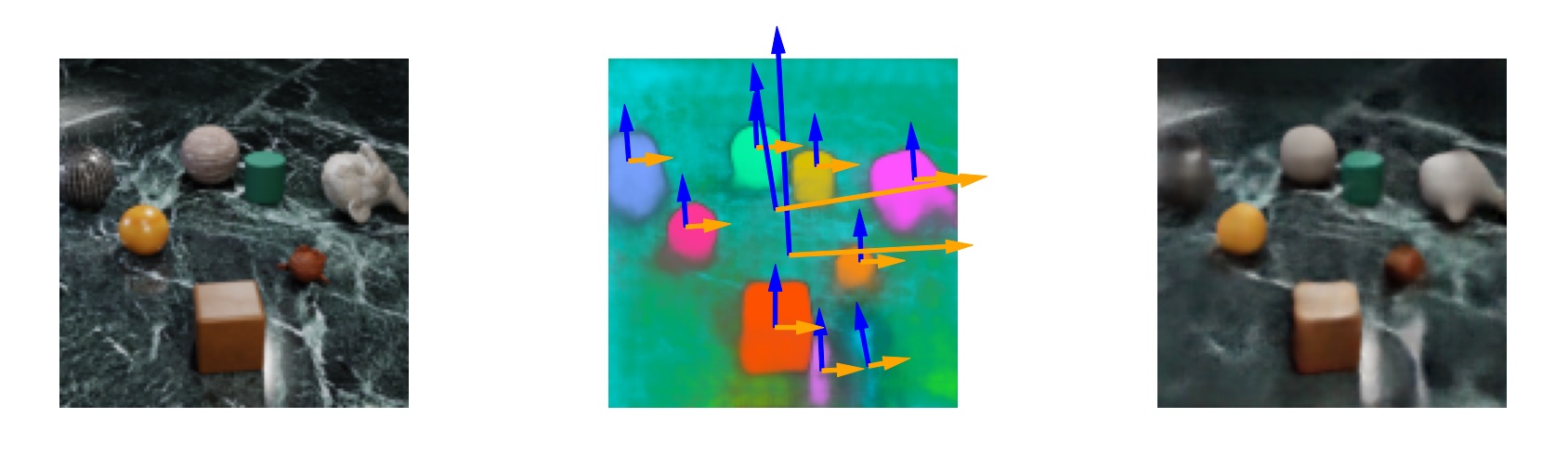}
    \includegraphics[height=\sizeframes]{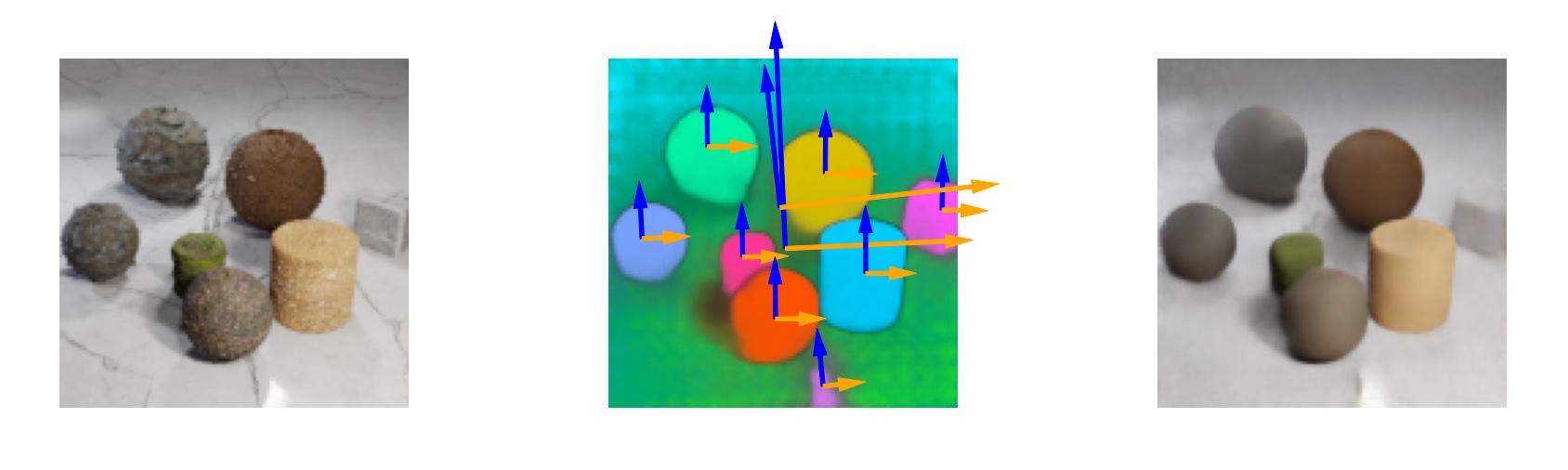}
    \caption{Per-slot reference frames learned without supervision on CLEVRTex. Left column: input image. Right column: reconstructed image. Middle column: we show the (x, y) position, (sx, sy) scale and orientation of individual slots overlaid over a predicted soft segmentation mask. Each pair of blue and orange arrows corresponds to one slot. The longer the arrows, the higher the scale of the slot is. We filter out slots that have a low predicted probability in the segmentation mask (i.e. they are probably inactive).}
    \label{fig:clevrtex_frames_fig}
\end{figure*}

\begin{figure*}[t]
    \centering
    \includegraphics[width=\textwidth]{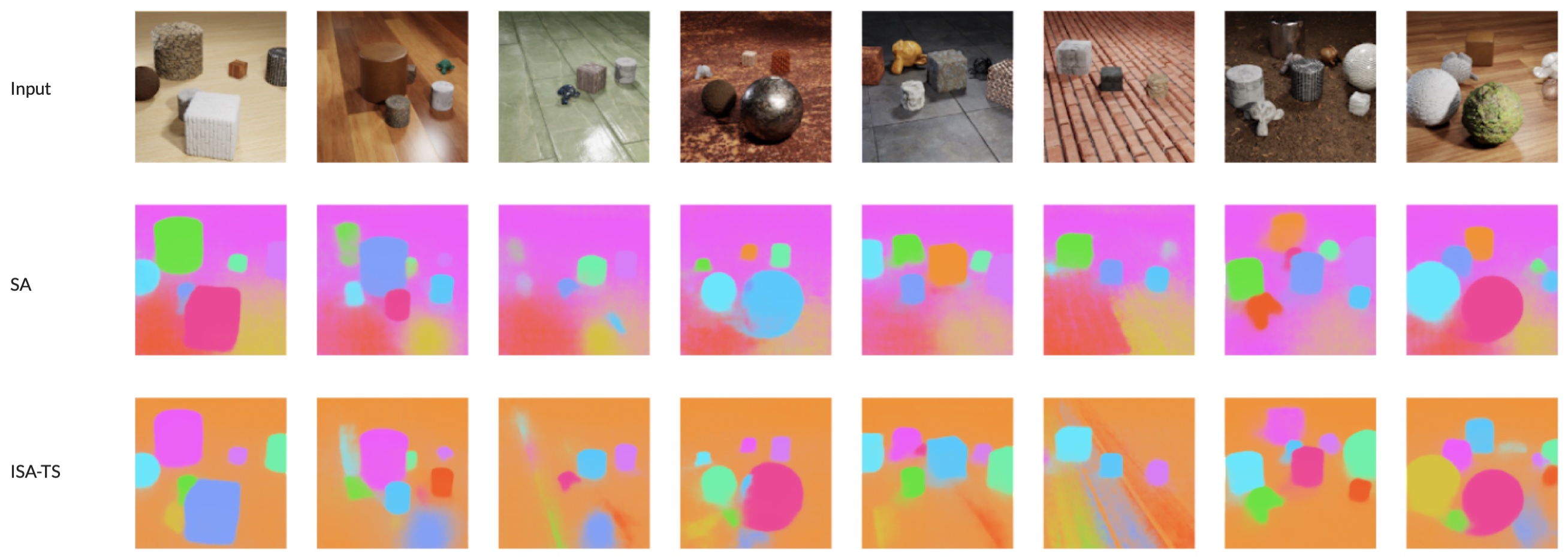}
    \caption{Comparison between Slot Attention and Translation and Scale Invariant Slot Attention in image segmentation.}
    \label{fig:clevrtex_segm}
\end{figure*}

\subsection{Waymo Open results}
Table \ref{tab:waymo} compiles all experiments pertaining to the Waymo Open dataset. 
\begin{table}[t]
    \centering
    \caption{Results on Waymo Open v1.4 with RGB and depth targets. All experiments used the ResNet-34 encoder and ran 10 random seeds.}
    \vspace{2mm}
    \begin{tabular}{llll}
        \toprule
        Method & Target & FG-ARI & MSE \\
        \midrule
        \begin{filecontents*}{csv/waymo_rgb.tex}
method,data,fg_ari_mean,fg_ari_std,mse_mean,mse_std
SA,RGB,27.6,12.7,584,35
\isats,RGB,\textbf{39.8},5.3,\textbf{523},12
- dec. only,RGB,32.4,7.1,586,29
\isatsr,RGB,31.2,8.4,561,26
\end{filecontents*}
\csvreader[
                column count=6,
                head to column names,
                late after line=\\
            ]{csv/waymo_rgb.tex}{1=\one, 2=\two, 3=\three, 4=\four, 5=\five, 6=\six}
            {\one & {\two} & {\three}{\color{lightgrey}\tiny$\pm$\four} & {\five}{\color{lightgrey}\tiny$\pm$\six}}
        \midrule
        \begin{filecontents*}{csv/waymo_depth.tex}
method,data,fg_ari_mean,fg_ari_std,mse_mean,mse_std
SA,Depth,40.6,5.0,\textbf{95},9
\isat,Depth,40.6,7.9,106,11
\isat-ABS,Depth,\textbf{45.2},12.2,96,9
\end{filecontents*}
\csvreader[
                column count=6,
                head to column names,
                late after line=\\
            ]{csv/waymo_depth.tex}{1=\one, 2=\two, 3=\three, 4=\four, 5=\five, 6=\six}
            {\one & {\two} & {\three}{\color{lightgrey}\tiny$\pm$\four} & {\five}{\color{lightgrey}\tiny$\pm$\six}}
        \bottomrule
    \end{tabular}
    \label{tab:waymo}
\end{table}

\end{document}